\documentclass[11pt,a4paper]{elsarticle}

\usepackage[utf8]{inputenc}
\usepackage[T1]{fontenc}

\usepackage{lmodern}
\usepackage{geometry}
\usepackage{graphicx}
\usepackage{subcaption}
\usepackage{amsmath, amssymb}
\usepackage{booktabs}
\usepackage{enumitem}
\usepackage{hyperref}
\usepackage{siunitx}
\usepackage{todonotes}
\usepackage{placeins}
\usepackage{textcomp}
\usepackage{lineno}

\usepackage{tikz}
\usetikzlibrary{calc,arrows.meta,positioning}

\usepackage{csquotes}

\graphicspath{{}{figs/}}

\journal{Journal of Computational Science}

\begin{document}

\begin{frontmatter}

\title{Transformation-driven generation of comparable projection images from multimodal anatomical scenes}

\author[a]{Dariusz Pojda}\corref{cor1}
\cortext[cor1]{Corresponding author, e--mail: dpojda@iitis.pl}
\author[a]{Krzysztof Domino}
\author[a]{Micha\l{} Tarnawski}
\author[a]{Agnieszka Anna Tomaka}

\affiliation[a]{
    organization={Institute of Theoretical and Applied Informatics, Polish Academy of Sciences},
    addressline={Ba\l{}tycka~5}, 
    postcode={44-100}, 
    city={Gliwice},
    country={Poland}}
    
\begin{abstract}

This work addresses the computational problem of generating reproducible projection-space observations from heterogeneous anatomical scenes whose components may undergo independent spatial transformations. We propose a transformation-driven framework for synthetic projection imaging from multimodal anatomical data and demonstrate it on mandibular-motion scenarios.

In contrast to conventional Digitally Reconstructed Radiograph (DRR) approaches primarily designed for registration, projection realism, or rendering efficiency, the proposed formulation treats projection imaging as an observation process operating on an explicitly represented anatomical scene.

Independently transformable volumetric and surface-based anatomical objects are embedded within a shared scene representation and propagated directly into projection space through explicit transformations. Projection geometry, acquisition modelling, material interpretation, and image presentation remain explicitly separated, enabling controlled exploration of methodological assumptions while preserving reproducibility and direct comparability between generated projections.

Particular emphasis is placed on transformation-driven anatomical scenarios relevant to craniofacial analysis, including mandibular motion and therapeutic repositioning. Using a shared anatomical reference scene composed of CT/CBCT volumes, segmented structures, surface models, and auxiliary anatomical or therapeutic objects, the framework enables generation of directly comparable VirtualRTG projections from multiple anatomical configurations while preserving identical imaging assumptions.

Rather than aiming at fully physically faithful radiographic simulation, the proposed approach provides a controllable and reproducible methodological environment for studying anatomy--projection relationships, motion observability, and transformation-aware imaging workflows.

\end{abstract}




\end{frontmatter}

\section{Introduction}

Advances in radiographic imaging have enabled observation of anatomical
structures during motion through dynamic acquisition techniques.
At the same time, increasing awareness of the harmful effects of ionising
radiation has reinforced the need to restrict imaging procedures to
clinically justified scenarios. In practice, dynamic radiographic imaging
remains additionally constrained by trade-offs between spatial and temporal
resolution, limited repeatability of motion conditions, and difficulties in
performing controlled methodological experiments.

These limitations motivate the development of synthetic projection imaging
methods capable of generating repeatable and controllable radiographic
scenarios without repeated exposure. Although virtual X-ray and
\emph{Digitally Reconstructed Radiograph} (DRR) methodologies have been
studied for decades, contemporary multimodal computational environments
introduce new requirements associated with heterogeneous anatomical data and
explicit modelling of anatomical motion.

Modern volumetric imaging techniques, particularly CT and CBCT, provide
high-resolution anatomical representations of craniofacial structures,
including skeletal anatomy, dentition, and selected soft tissues.
At the same time, segmentation, registration, and treatment-planning
workflows increasingly operate on multimodal anatomical scenes in which
volumetric datasets coexist with segmented surfaces, dental scans,
therapeutic appliances, and auxiliary geometric models. Such entities often
remain represented as independent objects possessing explicit spatial
relationships and transformation histories originating from registration,
motion tracking, biomechanical simulation, or treatment planning.

In contemporary multimodal anatomical workflows, the computational object of interest is often not a single volume but a scene composed of several spatially related entities, including CT or CBCT volumes, segmented anatomical components, optical surface scans, reconstructed meshes, landmarks, trajectories, and treatment-related geometries. Reducing such a scene to one resampled volume may simplify projection generation, but it also hides object-specific transformations, may introduce interpolation artefacts, and makes it difficult to analyse how individual anatomical changes affect the resulting projection. The proposed framework therefore keeps anatomical entities as independent scene objects and evaluates their current spatial state during projection generation.

A particularly relevant application concerns mandibular motion analysis, important in temporomandibular joint assessment, orthodontics, prosthodontics, craniofacial biomechanics, and therapeutic occlusal repositioning. Clinically relevant scenarios frequently involve predefined anatomical configurations, including protrusion, lateral motion, controlled mouth opening, or splint-related therapeutic repositioning. Repeated radiographic acquisition for such scenarios is often impractical due to radiation burden, limited repeatability of positioning, and restricted experimental control.

In such settings, projection imaging may be reformulated not as an isolated rendering or reconstruction problem, but as a controlled simulation of projection-space observations generated from explicitly represented anatomy and anatomical transformations embedded within a computational scene. This perspective motivates the concept of \emph{transformation-driven projection imaging}, in which synthetic projection images emerge directly from known scene geometry and object transformations. Consequently, a shared anatomical reference scene, derived from a single CT/CBCT acquisition, segmented volumetric components, or multiple registered volumetric and surface-based sources, may be used for systematic exploration of multiple anatomical configurations while preserving identical acquisition geometry and imaging assumptions.

\subsection{Contributions of this work}

In contrast to conventional DRR frameworks primarily designed for image registration, projection realism, or computational acceleration, the present work introduces a scene-oriented computational formulation of synthetic projection imaging. Projection images are treated as simulated observations of a transformation-aware anatomical scene rather than as isolated renderings of a single pre-resampled volume. As a result, projection-space representations emerge as deterministic consequences of explicitly represented anatomy, spatial relationships, and scene transformations.

The principal methodological contribution of this work is the formulation of
projection imaging as a transformation-aware process in which anatomical
motion and alternative anatomical configurations are propagated directly into
projection space through explicit scene transformations. Within this
formulation, anatomical structures are represented as independently
transformable volumetric and surface-based objects embedded in a shared
hierarchical scene model, enabling heterogeneous anatomical representations
to coexist while preserving explicit spatial relationships.

A further contribution lies in the explicit separation of anatomical
representation, projection geometry, acquisition modelling, material
interpretation, and presentation. Such decomposition enables controlled
methodological experimentation in which the influence of individual modelling
assumptions may be analysed independently without modifying the remaining
stages of the projection pipeline.

The framework additionally enables repeated generation of directly comparable
VirtualRTG projections for multiple anatomical configurations derived from a
shared anatomical reference scene. Such a scene may include a single CT/CBCT
dataset, segmented volumetric components, multiple registered volumetric
datasets, surface models, dental scans, therapeutic appliances, and auxiliary
geometric objects, while preserving identical acquisition geometry, material
assumptions, and presentation settings.

Such functionality is particularly relevant in scenarios where repeated
radiographic acquisition would be impractical or undesirable. By deriving
multiple projection images from a common anatomical reference scene, the
framework enables systematic investigation of projection-space effects induced
by anatomically defined transformations, such as mandibular motion, TMJ
repositioning, orthodontic treatment planning, or occlusal-splint-related
changes, without requiring additional patient exposure.

Finally, the proposed framework is implemented in Python within \texttt{pyDpVision}, a Python-based research environment derived from the dpVision platform \cite{pojda2025dpvision}, and is evaluated using synthetic and real anatomical projection scenarios, including multimodal scene integration, alternative projection backends, and transformation-aware motion experiments. The framework is intended primarily as a research and experimentation platform for studying relationships between anatomy, transformations, and projection-space observations. The design prioritises explicit scene representation, reproducibility of experimental conditions, and flexibility of anatomical transformations over maximisation of computational throughput. Consequently, computational performance is not the primary optimisation target of the present work.
The proposed projection framework extends the transformation-tree paradigm previously introduced for multimodal image registration and data integration \cite{TOMAKA2025110311} into the domain of synthetic projection imaging.

Importantly, the contribution of the present work does not lie in a new attenuation model or DRR rendering algorithm. Instead, it lies in the formulation of projection imaging as an observation process operating on a transformation-aware multimodal anatomical scene, where projection generation constitutes one component of a broader scene-management framework.

The computational-science role of the proposed framework can be summarised as a controlled simulation pipeline linking multimodal anatomical scene representation, transformation modelling, projection acquisition, and comparable projection-space observations (Table~\ref{tab:computational_science_framing}).

\begin{table}[!ht]
\centering
\caption{Computational-science interpretation of the proposed transformation-driven projection framework.}
\label{tab:computational_science_framing}
\small
\begin{tabular}{p{0.18\linewidth} p{0.72\linewidth}}
\toprule
Component & Role in the proposed framework \\
\midrule
Input & A multimodal anatomical scene composed of volumetric datasets, segmented structures, surface models, landmarks, trajectories, and auxiliary or treatment-related geometries. \\
Model & A transformation hierarchy combined with an explicit projection operator, allowing independently represented objects to preserve their spatial relationships and transformation histories. \\
Simulation & Controlled projection generation under configurable acquisition assumptions, including projection geometry, ray integration strategy, material-response model, depth limitation, and presentation settings. \\
Output & Directly comparable synthetic projection-space observations generated from alternative anatomical configurations while preserving identical imaging assumptions. \\
Use & Methodological analysis of anatomy--projection relationships, motion observability, transformation-aware validation datasets, planning scenarios, and controlled comparison of projection strategies. \\
\bottomrule
\end{tabular}
\end{table}

\section{Related Work}

Synthetic projection imaging and \emph{Digitally Reconstructed Radiographs} (DRRs) are commonly formulated as computational models of X-ray image formation based on attenuation of radiation along source--detector paths, typically described by the Beer--Lambert law~\cite{Lambert1760,Bushberg2012}:

\[
I = I_0 \exp\!\left(-\int \mu(s)\,\mathrm{d}s\right).
\]

In practice, synthetic projection generation reduces to numerical estimation
of attenuation along rays traversing an anatomical representation.
Depending on modelling assumptions, radiographic simulation may additionally
incorporate detector response, blur, noise, scattering, or secondary
interactions. However, many practical DRR frameworks adopt simplified
assumptions to maintain computational tractability.

For volumetric representations such as CT or CBCT, attenuation integration is
typically performed using voxel-traversal or interpolation-based strategies.
The \emph{Siddon} method~\cite{Siddon1985} computes exact voxel intersections,
whereas \emph{Jacobs}~\cite{Jacobs1998} introduces an incremental optimisation
of Siddon traversal. Alternatively, the \emph{Joseph}
approach~\cite{Joseph1982} approximates attenuation through sampled
interpolation along rays and remains widely used due to its favourable balance
between numerical accuracy and computational efficiency. Other formulations
include voxel-driven or footprint-based approaches for efficient projection
generation~\cite{Zollei2001,Westover1990,DeMan2004}, while Monte Carlo
frameworks such as \textit{Geant4}~\cite{Agostinelli2003} provide physically
realistic modelling of photon transport and detector response at substantially
higher computational cost. More advanced formulations additionally incorporate
wave-based effects, such as phase-contrast X-ray simulation~\cite{Peterzol2007}.

Beyond projection generation itself, DRR methodologies have been widely used
for 2D--3D registration between CT and fluoroscopic imagery, intraoperative
visualisation, panoramic reconstruction, and CBCT-derived cephalometric
analysis. Registration-oriented frameworks commonly employ DRRs as an
intermediate representation enabling alignment between volumetric anatomy and
projection-space observations~\cite{Zollei2001MICCAI,Hurvitz2008}. Increasing
clinical demand further motivated acceleration strategies for near real-time
projection generation~\cite{Galigekere2003}. In dentistry and
craniomaxillofacial imaging, CBCT-derived projections have been applied to
panoramic reconstruction, cephalometric analysis, and geometry-aware
projection synthesis~\cite{Papakosta2017,YUN2019205,electronics11152404,DBLP:conf/cimaging/LeeWLLS19},
while several studies demonstrated diagnostic comparability of CBCT-derived
cephalograms with conventional radiographs~\cite{Cattaneo2008,Kim2012,diagnostics11122292}.
Recent developments additionally explore dynamic imaging and differentiable
DRR formulations for motion analysis and inverse optimisation
problems~\cite{Seah2022,Gopalakrishnan2022}.

Despite substantial progress, several methodological limitations remain.

First, most existing DRR approaches are organised around one or more volumetric datasets and primarily target image registration, projection realism, tomographic reconstruction, or computational acceleration. In contrast, contemporary multimodal workflows frequently operate on heterogeneous anatomical scenes composed of volumetric datasets, segmented anatomical structures, surface models, treatment-related geometries, annotations, and motion-tracking outputs linked through explicit spatial relationships.

Second, existing methodologies predominantly address static anatomical configurations or short-term image-alignment tasks. Comparatively less attention has been devoted to frameworks supporting systematic generation of multiple directly comparable projection-space representations corresponding to explicitly defined anatomical transformations and motion scenarios, particularly in craniomaxillofacial applications involving mandibular kinematics, therapeutic repositioning, or occlusal splints.

Third, although physically realistic and differentiable DRR frameworks continue to evolve, relatively fewer studies focus on transparent and reproducible projection environments in which anatomy, transformations, projection geometry, acquisition assumptions, and presentation remain explicitly represented and independently controllable.

Contemporary frameworks such as Plastimatch \cite{plastimatch}, RTK \cite{rtk}, TIGRE \cite{tigre}, and DeepDRR \cite{deepDRR} represent important advances in DRR generation, image registration, tomographic reconstruction, and realistic radiographic simulation. Although these systems differ in implementation details and modelling assumptions, projection generation is typically formulated with respect to one or more volumetric datasets and a specified acquisition geometry.
While these frameworks provide powerful projection-generation capabilities, their primary focus lies on projection rendering, reconstruction, registration, or simulation rather than management of multimodal anatomical scenes represented as independently transformable objects.

To the best of our knowledge, none of these frameworks provides a general scene-oriented representation in which multiple independently transformable volumetric and surface-based anatomical objects can be combined, manipulated, and projected within a common transformation hierarchy while preserving direct comparability between alternative anatomical configurations.

The framework proposed in the present work adopts a different perspective. Rather than treating projection generation as a volume-centred operation, it treats projection images as observations of a transformation-aware anatomical scene composed of independently transformable volumetric and surface-based objects. Consequently, anatomical motion, multimodal integration, therapeutic configurations, anatomical annotations, and alternative projection scenarios are represented within a common scene model and propagated directly into projection space through explicit transformations.

The primary objective is therefore not maximisation of projection realism or computational performance, but support for controlled generation of directly comparable projection images from complex multimodal anatomical scenes. The present work addresses the above limitations through a transformation-driven projection-imaging framework supporting independently transformable multimodal anatomical objects, repeated generation of comparable projections for controlled anatomical configurations, and explicit integration of motion-related scenarios, with particular emphasis on mandibular movement and therapy-related positioning.
From this perspective, the primary objective is not merely projection
generation itself, but the creation of a controlled experimental environment
for studying how explicitly defined anatomical transformations manifest in
projection space.

\section{Method}

The methodological objective of the proposed framework is generation of
multiple anatomically consistent synthetic projection images from a shared
anatomical reference scene under controlled anatomical transformations. Rather
than treating projection generation as an isolated rendering problem, the
proposed approach formulates image formation as a downstream consequence of
explicitly represented anatomy, projection geometry, acquisition assumptions,
and presentation rules.

In contrast to approaches requiring repeated acquisition or reconstruction of
separate volumetric datasets for alternative imaging scenarios, the proposed
framework assumes repeated reuse of a shared anatomical scene. Depending on the
application, this scene may consist of a single CT/CBCT volume, segmented
volumetric components derived from one acquisition, or multiple registered
volumetric and surface-based models originating from different sources.
Anatomical configurations are then modified through explicitly represented
spatial transformations. Clinically relevant examples include therapeutic
mandibular repositioning, occlusal splints, protrusion, lateral motion, or
controlled mouth opening.

For clarity, we summarise the proposed decomposition using the following
abstract notation:
\[
\mathcal{I}_{\mathrm{display}} =
\mathcal{P}\!\left(
\mathcal{A}\!\left(
\mathcal{S},
\mathcal{G},
\Theta_A
\right),
\Theta_P
\right),
\]
where \(\mathcal{S}\) denotes an anatomical scene composed of independently
transformable objects and their spatial relationships, \(\mathcal{G}\) defines
projection geometry, \(\Theta_A\) represents acquisition parameters, including
sampling, attenuation modelling, material interpretation, and ray integration,
while \(\Theta_P\) denotes presentation parameters responsible for the final
visual appearance of the image.

The operator \(\mathcal{A}\) corresponds to synthetic projection acquisition,
whereas \(\mathcal{P}\) represents post-acquisition presentation and visual
interpretation of the generated signal. This distinction explicitly separates
projection-space information resulting from anatomical and acquisition
assumptions from display-related choices such as contrast, inversion, tonal
mapping, or display windowing.

This notation extends the standard DRR image-formation view, in which
attenuation is accumulated along source--detector rays and converted into image
intensity, by explicitly separating scene representation, acquisition
assumptions, and presentation.

Because transformation-driven imaging assumes repeated generation of comparable
projections under controlled anatomical motion and heterogeneous multimodal data,
the framework adopts an explicit methodological decomposition. In particular,
the method separates anatomical scene representation and spatial reference
modelling, projection geometry, projection acquisition and attenuation
modelling, and presentation from one another. Such separation enables independent
analysis of the influence of anatomical transformations, acquisition geometry,
material assumptions, and presentation strategy while preserving repeated reuse
of the same anatomical scene across multiple controlled imaging experiments.

\subsection{Methodological assumptions and computational environment}

The proposed framework assumes that anatomical structures remain represented as independent scene objects throughout the projection workflow. Rather than being merged into a single resampled volume, anatomical components such as bones, teeth, soft tissues, therapeutic appliances, and auxiliary geometric models preserve their own coordinate systems and spatial transformations.

This assumption reflects the organisation of multimodal computational environments used in registration, treatment planning, and motion analysis, where heterogeneous anatomical representations coexist and are related through explicit transformations.

Consequently, synthetic projection generation is formulated as a transformation-aware observation of an anatomical scene rather than as a projection of a pre-computed unified volume. Alternative anatomical configurations are represented through modifications of scene transformations, while projection geometry and imaging parameters remain independently configurable.

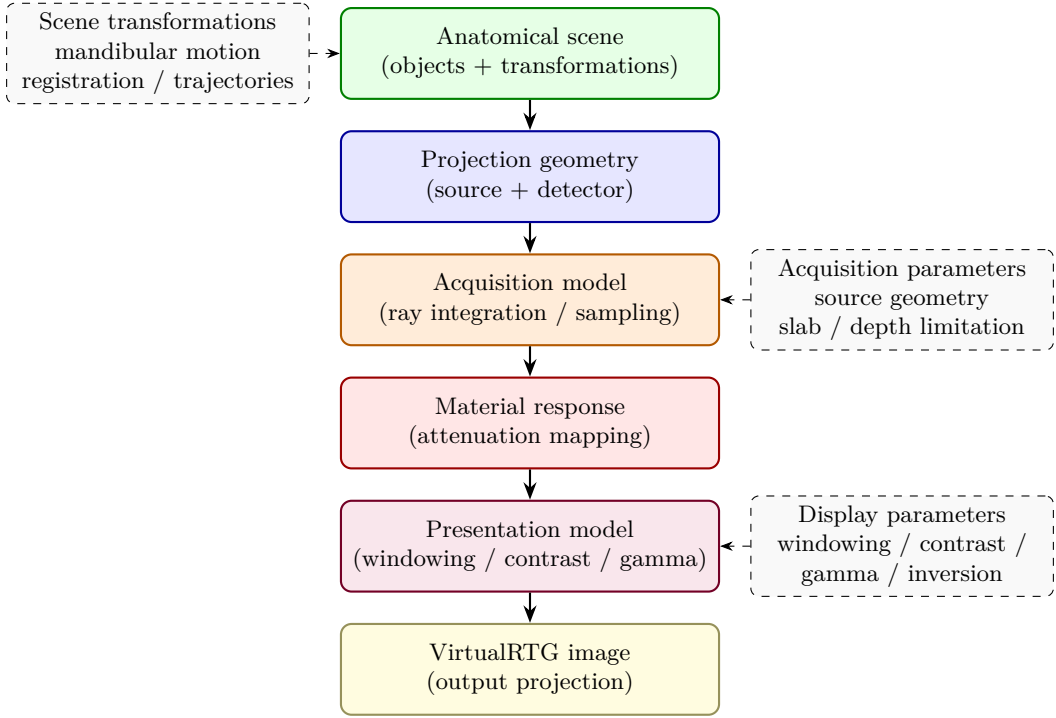
\begin{figure}[!h]
\centering

\begin{tikzpicture}[
    >=Stealth,
    font=\footnotesize,
    block/.style={
        rounded corners=4pt,
        draw,
        thick,
        minimum width=5.0cm,
        minimum height=1.2cm,
        align=center
    },
    param/.style={
        rounded corners=3pt,
        draw,
        dashed,
        fill=gray!5,
        minimum width=4.0cm,
        minimum height=0.9cm,
        align=center
    }
]


\node[
block,
fill=green!10,
draw=green!50!black
]
(scene)
{Anatomical scene\\
(objects + transformations)};

\node[
block,
fill=blue!10,
draw=blue!60!black,
below=0.4cm of scene
]
(geometry)
{Projection geometry\\
(source + detector)};

\node[
block,
fill=orange!15,
draw=orange!70!black,
below=0.4cm of geometry
]
(acquisition)
{Acquisition model\\
(ray integration / sampling)};

\node[
block,
fill=red!10,
draw=red!60!black,
below=0.4cm of acquisition
]
(material)
{Material response\\
(attenuation mapping)};

\node[
block,
fill=purple!10,
draw=purple!60!black,
below=0.4cm of material
]
(presentation)
{Presentation model\\
(windowing / contrast / gamma)};

\node[
block,
fill=yellow!15,
draw=yellow!50!black,
below=0.4cm of presentation
]
(result)
{VirtualRTG image\\
(output projection)};


\draw[->, thick] (scene) -- (geometry);
\draw[->, thick] (geometry) -- (acquisition);
\draw[->, thick] (acquisition) -- (material);
\draw[->, thick] (material) -- (presentation);
\draw[->, thick] (presentation) -- (result);


\node[
param,
left=0.4cm of scene
]
(transforms)
{Scene transformations\\
mandibular motion\\
registration / trajectories};

\node[
param,
right=0.4cm of acquisition
]
(physics)
{Acquisition parameters\\
source geometry\\
slab / depth limitation};

\node[
param,
right=0.4cm of presentation
]
(vis)
{Display parameters\\
windowing / contrast /\\
gamma / inversion};


\draw[->, dashed]
(transforms.east)
-- (scene.west);

\draw[->, dashed]
(physics.west)
-- (acquisition.east);

\draw[->, dashed]
(vis.west)
-- (presentation.east);

\end{tikzpicture}

\caption{
Methodological decomposition of the proposed projection framework. Synthetic projection generation is organised into explicit stages comprising anatomical scene representation, projection geometry, acquisition, material-response modelling, and presentation.
Separation of these stages enables independent experimentation with imaging geometry, anatomical transformations, material assumptions, and presentation strategies while preserving a common anatomical scene representation.
}
\label{fig:methodological_decomposition}
\end{figure}

The framework follows a modular architecture comprising scene representation, projection geometry, acquisition, material-response modelling, and image presentation (Fig.~\ref{fig:methodological_decomposition}). Separation of these stages enables independent experimentation with anatomical transformations, acquisition assumptions, attenuation models, and visualisation strategies while preserving a common anatomical reference.

An important consequence of this formulation is the absence of mandatory resampling into a common projection volume. During projection generation, rays are evaluated directly in the current transformation state of individual scene objects, allowing independently transformable anatomical structures to participate in image formation without modification of the original data.

The resulting organisation provides a reproducible environment for studying the relationships between anatomy, motion, imaging geometry, and projection-space representation under controlled conditions.

\subsection{Scene representation and coordinate transformations}

Anatomical objects participating in projection generation are represented
as independent scene entities with their own coordinate systems and
transformations. Consequently, projection generation does not require
construction of a globally resampled anatomical volume. Instead, object
positions are evaluated dynamically through transformation propagation
during ray integration.

The transformation hierarchy used throughout the framework follows the scene-management principles previously introduced in the transformation-tree model proposed for multimodal image registration and data integration \cite{TOMAKA2025110311}.

Projection sampling is formulated as a sequence of coordinate
transformations
\[
\mathbf{p}_{\mathrm{ref}}
\rightarrow
\mathbf{p}_{\mathrm{world}}
\rightarrow
\mathbf{p}_{\mathrm{local}}
\rightarrow
\mathbf{p}_{\mathrm{voxel}}
\]
allowing anatomical objects originating from volumetric or surface-based representations to participate in multiple spatial configurations without modification of their underlying data.

Beyond CT and CBCT volumes, the framework supports surface-based
representations described by triangular meshes. Volumetric and surface
objects participate in the same transformation hierarchy, enabling
construction of multimodal anatomical scenes integrating volumes,
segmented structures, dentition models, therapeutic appliances, and
auxiliary geometries.

The same transformation chain may also be applied to anatomical
annotations. Consequently, landmarks and paths remain spatially linked
to anatomy and may be propagated consistently into projection space.
This enables direct correspondence between three-dimensional anatomical
representations and their projection-space visualisations.

\subsection{Projection geometry model}

Projection geometry is represented independently from the anatomical scene.
Consequently, the same anatomical configuration may be projected under
different acquisition geometries, while identical geometry may be preserved
for comparison of multiple anatomical configurations.

The detector is defined by its origin $\mathbf{o}_D$, horizontal and vertical
pixel-step vectors $\mathbf{u}_D$ and $\mathbf{v}_D$, and image resolution
$H\times W$.

The centre of detector pixel $(r,c)$ is given by

\[
\mathbf{p}_D(r,c)=
\mathbf{o}_D
+
\left(c+\frac12\right)\mathbf{u}_D
+
\left(r+\frac12\right)\mathbf{v}_D.
\]

The framework supports both cone-beam and parallel projection models.
For cone-beam projection, rays originate from a point source
$\mathbf{s}$ and pass through detector pixels according to

\[
\hat{\mathbf d}(r,c)=
\frac{
\mathbf{p}_D(r,c)-\mathbf{s}
}{
\left\|
\mathbf{p}_D(r,c)-\mathbf{s}
\right\|
}.
\]

In parallel projection, all rays share a common direction,
eliminating perspective effects and facilitating comparative
and methodological studies.

This formulation enables systematic exploration of the influence
of acquisition geometry while preserving a common anatomical
reference model.

\subsection{Volumetric integration and acquisition model}

Following projection-geometry definition, projection generation requires
numerical evaluation of attenuation along rays traversing the anatomical
scene. In the proposed framework, this acquisition stage may combine
volumetric sources, surface-based anatomical structures, and optional
depth-limiting constraints within a single transformation-aware projection
process.

The following subsections describe volumetric integration, surface-based
projection, attenuation accumulation, detector response, and depth-limited
projection.

\subsubsection{Volumetric integration backends}

The framework supports two volumetric integration strategies: interpolated ray sampling inspired by the Joseph approach~\cite{Joseph1982} and exact voxel traversal based on the Siddon algorithm~\cite{Siddon1985}. Both strategies evaluate attenuation along projection rays, but they differ in how the ray--volume interaction is discretised.

In the Joseph-type sampling model, each projection ray is uniformly sampled along its path. For a ray originating at \(\mathbf{o}_R\) and travelling in direction \(\hat{\mathbf d}\), sampling locations are defined as
\[
\mathbf p(t_k)
=
\mathbf o_R+t_k\hat{\mathbf d},
\qquad
t_k=k\Delta s,
\]
where \(\Delta s\) denotes the ray-marching step size. Local attenuation values are reconstructed by interpolation of the volumetric data at intermediate positions. In practice, trilinear interpolation provides a favourable compromise between projection smoothness, suppression of staircase artefacts, and computational efficiency.

The Siddon-based formulation instead partitions the ray according to exact intersections with voxel-boundary planes. The ordered traversal parameters
\[
\tau_0 < \tau_1 < \cdots < \tau_M
\]
define segments fully contained within individual voxels, leading to the line integral
\[
\mathcal I
=
\sum_{m=0}^{M-1}
\mu(\mathbf{i}_m)\,\ell_m,
\]
where \(\mathbf{i}_m\) denotes the traversed voxel and \(\ell_m=\tau_{m+1}-\tau_m\) is the corresponding chord length. This formulation is independent of the sampling step \(\Delta s\), but assumes piecewise homogeneous attenuation within voxel boundaries.

\begin{figure}[!h]
\centering
\includegraphics[width=\textwidth]{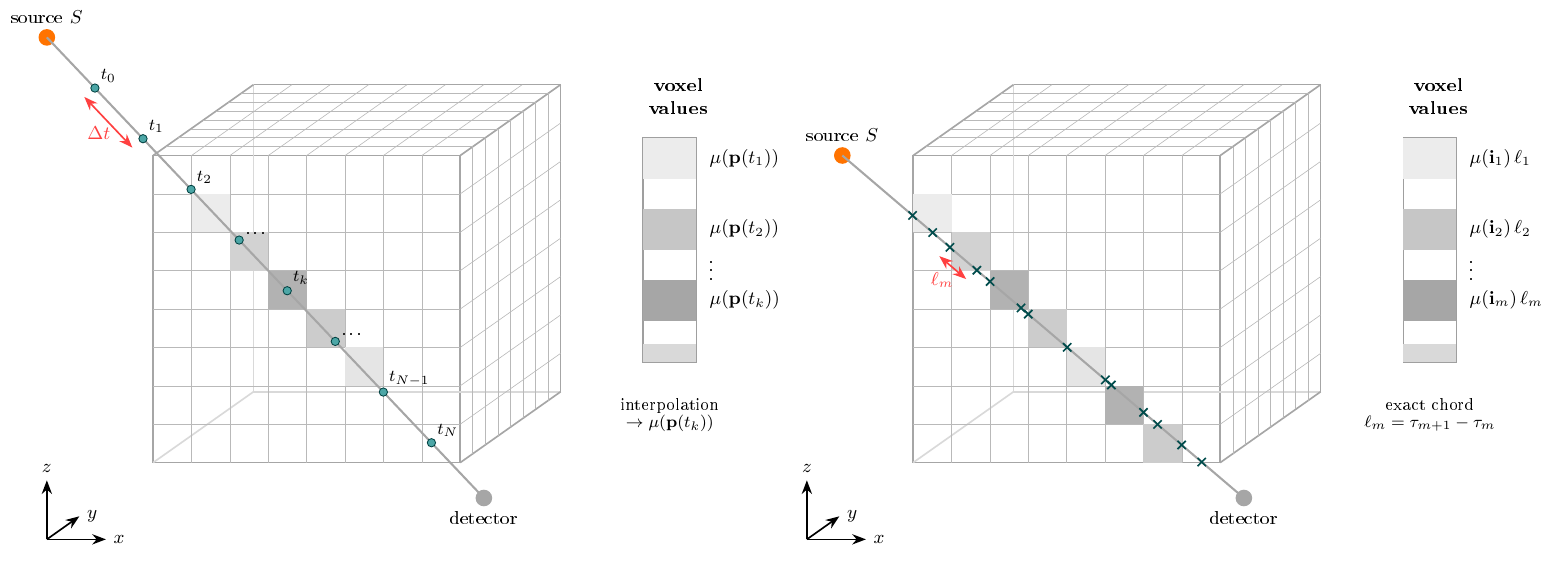}
\caption{
Comparison of volumetric integration strategies.
Left: Joseph-type sampling, in which a projection ray is discretised into sampling points separated by step \(\Delta s\) and attenuation values are reconstructed through interpolation.
Right: Siddon exact voxel traversal, where ray--voxel intersections partition the ray into segments fully contained within individual voxels, each contributing proportionally to its chord length \(\ell_m\).
}
\label{fig:integration_backends}
\end{figure}

Figure~\ref{fig:integration_backends} illustrates the difference between both integration strategies. The Joseph-type model provides flexibility and favourable computational efficiency, whereas the Siddon formulation serves as a step-size-independent reference in scenarios requiring exact voxel traversal.

\subsubsection{Surface-based projection}

Besides volumetric attenuation sources, the framework supports anatomical
structures represented as triangular surface meshes. Such objects participate
in the same transformation-aware projection pipeline as volumetric data, but
their contribution is evaluated through ray--surface intersection rather than
voxel sampling.

Two surface-projection strategies are considered. The first one follows a
ray-based formulation, in which each projection ray is tested against a
bounding-volume hierarchy and the resulting ray--triangle intersections are
accumulated along the ray path. The second one follows a detector-space
formulation, in which surface triangles are projected onto the detector and
stored as per-pixel intersection lists prior to attenuation accumulation.

The ray-based formulation is conceptually direct and preserves a close
geometric interpretation of ray--mesh intersections. The detector-space
formulation shifts part of the computation to a preprocessing stage and may
therefore be more efficient for selected projection scenarios. In both cases,
surface models remain independently transformable scene objects and may be
combined with volumetric data within the same projection process.

\subsubsection{Signal accumulation and detector response}

Given local attenuation responses provided by the material-response model,
projection formation is performed by accumulating attenuation contributions
along each projection ray. If the scene contains \(M\) active anatomical or
auxiliary objects, the total attenuation response at sampling location \(t_k\)
is expressed as
\[
\mu_{\mathrm{tot}}(t_k)
=
\sum_{m=1}^{M}
\mu_m(t_k),
\]
where \(\mu_m(t_k)\) denotes the contribution of the \(m\)-th object.

For the sampling-based backend, the accumulated attenuation is approximated as
\[
A
\approx
\sum_{k=0}^{N-1}
\mu_{\mathrm{tot}}(t_k)\,
\Delta s,
\]
where \(N\) denotes the number of sampling points and \(\Delta s\) is the
ray-marching step. For the Siddon backend, the accumulation follows the
voxel-traversal formulation introduced in the previous subsection and remains
independent of \(\Delta s\).

The accumulated attenuation may either be used directly,
\[
I_{\mathrm{raw}} = A,
\]
or converted into detector intensity using a simplified Beer--Lambert-type
model,
\[
I_{\mathrm{raw}}
=
\max\!\left(
I_{\mathrm{floor}},
e^{-A}
\right),
\]
where \(I_{\mathrm{floor}}\) defines a lower signal bound for numerical
stability.

This formulation intentionally omits scattering, polychromatic source spectra,
detector-specific response, and quantum noise. The aim is therefore not a fully
physical simulation of radiographic acquisition, but a repeatable and stable
projection model suitable for comparing anatomical configurations and
projection geometries under controlled assumptions.

For cone-beam geometries, an optional distance-dependent correction may be
applied as a simplified geometric intensity factor,
\[
G(d)=
\left(
\frac{d_{\mathrm{ref}}}{d}
\right)^p,
\]
where \(d\) is the source--detector distance, \(d_{\mathrm{ref}}\) is a
reference distance, and \(p\) controls the correction strength.

\subsubsection{Depth-limited projection (slab)}

Instead of integrating attenuation over the full ray extent, projection
formation may optionally be restricted to a selected spatial interval. In the
sampling-based formulation, this corresponds to limiting attenuation
accumulation to
\[
t\in[t_{\min},t_{\max}],
\]
which leads to
\[
A
\approx
\sum_{k=k_{\min}}^{k_{\max}}
\mu_{\mathrm{tot}}(t_k)\,
\Delta s.
\]

Two depth-limiting strategies are supported. In the \texttt{ray} mode, the
active interval is defined directly along each projection ray. In the
\texttt{planar} mode, attenuation accumulation is restricted to a slab
orthogonal to axis \(\hat{\mathbf n}\), such that a point \(\mathbf p\)
contributes only if
\[
s_{\min}
\le
(\mathbf p-\mathbf o)\cdot\hat{\mathbf n}
\le
s_{\max}.
\]

\begin{figure}[!h]
\centering

\begin{tikzpicture}[>=Stealth, font=\small,
slab/.style={fill=blue!10, draw=blue!40, opacity=0.7},
ray/.style ={->, thick, gray!70},
dim/.style ={<->, thin, gray},
]

\coordinate (S) at (0,4);
\fill[orange!80!red] (S) circle (4pt)
node[above=3pt] {source $S$};

\draw[thick, black!50] (-3,0) -- (3,0)
node[right] {detector};

\fill[slab] (-3.5,1.4) rectangle (3.5,2.6);

\draw[blue!50, dashed]
(-3.5,1.4) -- (3.5,1.4)
node[right, blue!70] {$s_{\min}$};

\draw[blue!50, dashed]
(-3.5,2.6) -- (3.5,2.6)
node[right, blue!70] {$s_{\max}$};

\node[blue!70] at (-2.5,2.3)
{active slab};

\draw[teal, dotted] (0,0) -- (0,4);

\draw[->, very thick, teal]
(0,2.0)
-- node[right=2pt] {$\hat{\mathbf{n}}$}
(0,3.2);

\fill[teal!60] (0,2.0) circle (3pt)
node[right=2pt] {$\mathbf{o}$};

\foreach \x in {-2,-0.5,1.5}{
\draw[ray] (S) -- (\x,0);
}

\coordinate (P1) at ($(S)!0.35!(1.5,0)$);
\coordinate (P2) at ($(S)!0.65!(1.5,0)$);

\draw[very thick, red!70]
(P1) -- (P2);

\node[red!70, right=2pt]
at ($(P1)!0.5!(P2)$)
{$t\in[t_{\min},t_{\max}]$};

\draw[dim]
(-4,1.4)
--
(-4,2.6)
node[midway,left=3pt]
{$s_{\max}-s_{\min}$};

\end{tikzpicture}

\caption{
Depth-limited projection in the planar mode.
Only attenuation contributions originating from the selected slab interval
participate in projection formation.
}
\label{fig:slab-geometry}
\end{figure}
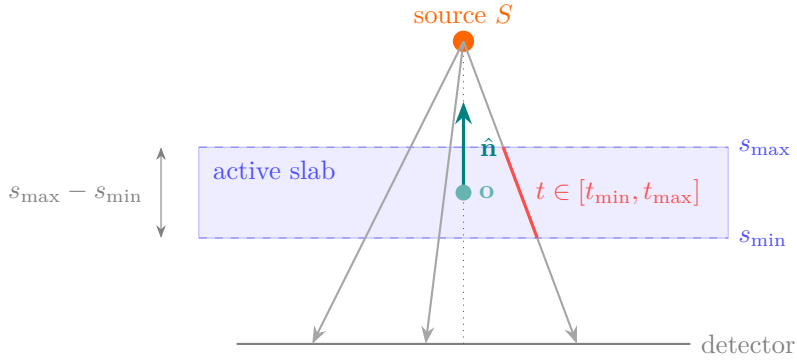

Figure~\ref{fig:slab-geometry} illustrates the planar slab formulation. By
restricting attenuation accumulation to a selected depth range, the mechanism
may reduce structural superposition and improve observability of anatomically
relevant regions.

\subsection{Material response model}

Projection formation requires a mapping between volumetric scalar values and
local attenuation response. In the proposed framework, this relation is treated
as a configurable material-response model rather than as a strict physical
simulation of X-ray propagation.

This assumption is particularly important for CBCT data, where voxel values do
not represent directly calibrated attenuation coefficients and may depend on
the acquisition protocol, reconstruction procedure, and scanner-specific
processing. Consequently, the framework separates scalar preprocessing,
scalar-to-attenuation mapping, and optional material-range weighting.

Prior to attenuation modelling, scalar values may be normalised to reduce
dataset-dependent intensity-range variability. This preprocessing step does not
constitute physical calibration; it only stabilises the subsequent
interpretation of scalar values across datasets.

After preprocessing, local attenuation is defined by a mapping
\[
I \mapsto \mu(I),
\]
where \(I\) denotes the volumetric scalar intensity and \(\mu(I)\) is the
corresponding local attenuation response. Several mappings may be used,
including a linear reference model and piecewise-linear formulations designed
to emphasise selected anatomical components, such as mineralised structures or
soft tissues.

For applications requiring smooth enhancement of bone-like intensities, a
soft-threshold formulation may be used. The contribution of mineralised
structures is controlled by a logistic weighting function,
\[
w(I)
=
\frac{1}{
1+\exp\!\left(
-\frac{I-T_b}{\sigma_b}
\right)
},
\]
where \(T_b\) denotes the preferred transition level and \(\sigma_b\) controls
transition smoothness. This avoids abrupt binary thresholding and provides a
continuous transition between weakly and strongly attenuating structures.

The framework also supports material windowing, which modifies attenuation
before ray integration. Unlike conventional display windowing, which affects
only the final visual appearance, material windowing changes the contribution
of selected scalar ranges to the projection formation process:
\[
\mu_{\mathrm{final}}(I)
=
\mu(I)\,
w_{\mathrm{window}}(I).
\]
This mechanism enables selective enhancement or suppression of anatomical
components and supports controlled analysis of structure observability.

All material-response models used in this work should be interpreted as
configurable methodological assumptions rather than physically calibrated
descriptions of X-ray interaction with tissue. Their purpose is to provide a
repeatable and controllable way to analyse how different anatomical structures
contribute to synthetic projection images.

\subsection{Presentation model}

An important characteristic of the proposed framework is the explicit
separation between projection acquisition and image presentation. The raw
projection signal represents the result of the acquisition and attenuation
model, whereas the displayed image is obtained through a separate
presentation stage. This distinction allows contrast, intensity range, tonal
mapping, and inversion parameters to be modified without recomputing the
underlying ray integration.

In general, presentation may be expressed as
\[
I_{\mathrm{display}}
=
P(I_{\mathrm{raw}};\Theta_P),
\]
where \(\Theta_P\) denotes display-related parameters such as intensity
window, contrast, gamma correction, and optional inversion. These parameters
influence only the visual appearance of the generated projection and do not
modify the underlying attenuation accumulation.

This separation is important for comparative experiments, because multiple
anatomical configurations may be evaluated under identical acquisition
assumptions while presentation settings are adjusted independently for
visual inspection.

\subsection{Computational considerations and quality profiles}

The computational cost of synthetic projection generation depends primarily
on detector resolution, the number of sampling steps along each ray, and the
number of anatomical objects contributing to projection formation.

For the sampling-based backend, the computational complexity may be
approximated as
\[
T
\sim
H\cdot W
\cdot
\frac{L}{\Delta s}
\cdot
N_v,
\]
where \(H\times W\) denotes detector resolution, \(L\) is the average ray
traversal length through the anatomical scene, \(\Delta s\) is the sampling
step, and \(N_v\) corresponds to the number of active volumetric objects.

Consequently, increasing detector resolution or reducing the sampling step
improves projection detail at the cost of increased computation time. To
support controlled trade-offs between computational cost and image fidelity,
the framework provides predefined quality profiles corresponding to different
detector resolutions and sampling densities.

The Siddon backend exhibits different scaling behaviour because the number of
integration segments depends on voxel-boundary traversal rather than uniform
ray discretisation. Nevertheless, both integration strategies expose a similar
practical trade-off between projection fidelity and execution time.

\section{Transformation-driven projection imaging}

The projection framework introduced in the previous sections enables
generation of synthetic radiographic images for a given anatomical
configuration and acquisition geometry. A central methodological extension
proposed in this work is the introduction of
\emph{transformation-driven projection imaging}, in which changes in
anatomical configuration are propagated directly into projection space through
explicit scene transformations.

Unlike conventional dynamic imaging workflows relying on repeated acquisition,
reconstruction of multiple volumetric datasets, or motion estimation directly
in image space, the proposed framework represents anatomy as a collection of
independently transformable objects embedded within a hierarchical scene
representation. Projection formation therefore becomes a deterministic
consequence of scene configuration rather than an isolated rendering task.

In the proposed formulation, anatomical motion is represented through explicit
transformations applied to anatomical objects participating in projection
formation. Since anatomical structures remain embedded in the same scene graph throughout
the workflow, alternative anatomical configurations may be explored without
modifying the underlying volumetric or surface-based data and without repeating
acquisition.

From a methodological perspective, mandibular motion constitutes a particularly
natural use case. Clinically relevant configurations, including therapeutic
repositioning, protrusion, lateral movement, or controlled mouth opening, may be represented as explicit transformations acting on a shared anatomical scene, for example a segmented CT/CBCT volume with independently transformable
mandibular and cranial components, or a multimodal scene combining volumetric
and surface-based anatomical representations.
Consequently, multiple directly
comparable VirtualRTG projections may be generated while preserving identical
projection geometry, acquisition assumptions, and presentation parameters.

\subsection{Transformation-aware motion representation}

Within the proposed framework, motion is represented as an explicit change of scene configuration. Anatomical structures remain independent scene objects participating in the same hierarchical transformation model used throughout the \texttt{dpVision}/\texttt{pyDpVision} environment.

A dynamic anatomical scene is therefore expressed as a time-dependent scene:
\[
\mathcal{S}(t),
\]
whose configuration evolves through transformations acting on selected anatomical objects.

For rigid anatomical motion, such as mandibular displacement, a natural representation is given by a rigid transformation belonging to the \(SE(3)\) group:
\[
T(t)=
\begin{bmatrix}
R(t) & \mathbf{t}(t) \\
0 & 1
\end{bmatrix},
\]
where \(R(t)\in SO(3)\) describes rotation, \(\mathbf{t}(t)\in\mathbb{R}^3\) translation, and \(t\) denotes either physical time or a configuration index.

Under this formulation, synthetic projection generation becomes explicitly dependent on anatomical configuration:
\[
\mathcal{I}_{\mathrm{display}}(t)=
\mathcal{P}
\!\left(
\mathcal{A}
\!\left(
\mathcal{S}(t),
\mathcal{G},
\Theta_A
\right),
\Theta_P
\right),
\]
where \(\mathcal{S}(t)\) denotes the current anatomical scene, \(\mathcal{G}\) defines projection geometry, \(\Theta_A\) represents acquisition parameters, and \(\Theta_P\) corresponds to presentation settings.

In the simplest formulation, projection geometry remains fixed and only the anatomical configuration evolves over time. However, the formalism naturally admits scenarios involving dynamic source or detector trajectories.

\subsection{Transformation-driven projection scenarios}

The proposed formulation supports both discrete anatomical configurations and
continuous motion trajectories.

Discrete scenarios correspond to explicitly defined anatomical states, including
reference position, therapeutic repositioning, protrusion, lateral movement, or
different degrees of mouth opening. In such cases, the starting point is a
shared anatomical reference scene, which may consist of a single CT/CBCT
volume, segmented volumetric components, multiple registered volumetric
datasets, surface models, or auxiliary therapeutic and geometric objects. The
scene is then modified through explicit anatomical transformations, while
projection geometry and acquisition assumptions may remain unchanged. Under
this interpretation, projection imaging becomes a mapping:
\[
\begin{gathered}
\text{shared anatomical reference scene}
\\[2mm]
\downarrow
\\[2mm]
\text{multiple anatomical configurations}
\\[2mm]
\downarrow
\\[2mm]
\text{multiple comparable VirtualRTG projections}
\end{gathered}
\]
Alternatively, continuous scenarios may be represented through trajectories:
\[
T(t)\in SE(3),
\]
allowing generation of temporally ordered synthetic projection sequences that
describe anatomical motion.

Because projection geometry, attenuation modelling, and presentation may remain
unchanged across all configurations, differences between generated images may be
interpreted primarily as consequences of anatomical transformations rather than
acquisition variability.

\subsection{Methodological implications}

Transformation-driven projection imaging provides a controlled computational environment in which anatomy, motion, projection geometry, and image formation remain explicitly represented and independently controllable.

This formulation enables systematic investigation of anatomy--projection relationships under controlled conditions and may support analysis of projection-space observability of anatomical motion, validation of motion estimation methods, generation of reproducible reference datasets, and development of machine-learning pipelines operating on projection data.

A particularly important implication concerns craniomaxillofacial analysis and mandibular motion. A shared anatomical reference scene may serve as a common basis for multiple
therapeutic or functional configurations. In the simplest case, this scene may
be derived from a single CT or CBCT acquisition; in more complex scenarios, it
may combine segmented volumetric components, multiple registered CBCT volumes,
surface scans, reconstructed anatomical models, and therapeutic objects.

The proposed formulation should therefore be interpreted not as a replacement for clinical fluoroscopy, but as a reproducible and transformation-aware experimental framework enabling controlled exploration of anatomical motion in projection space.

\section{Computational evaluation and transformation-driven use cases}\label{sec}

The evaluation was organised around four methodological questions:
(i) what computational cost is associated with the proposed projection backends and scene configurations;
(ii) how sampling-based and exact voxel-traversal backends differ in accuracy and performance;
(iii) whether heterogeneous anatomical objects can be projected within a common transformation-aware scene without mandatory global resampling; and
(iv) whether transformation-driven and depth-limited projection scenarios can improve controlled analysis of motion-dependent anatomical relationships.

The aim of this evaluation is not to establish diagnostic equivalence with clinical radiographs, but to assess whether the proposed computational formulation supports reproducible, comparable, and transformation-aware projection experiments involving volumetric, surface-based, and multimodal anatomical representations.

\subsection{Synthetic benchmark scenes and projection backend evaluation}

All measurements were performed on purely synthetic, in-memory datasets. No disk I/O or GUI rendering is included in the reported timings. The benchmark covers three quality profiles, three volume sizes, and three scene variants (volumetric only, mesh only, and hybrid).

Unless stated otherwise, all benchmarks were performed on a desktop workstation equipped with an Intel\textregistered{} Core\texttrademark{} Ultra 5 225 processor and 64 GB RAM (3200 MT/s), running Windows 11 Pro. No dedicated GPU acceleration was used in the reported experiments.

\paragraph{Volumes}
Three hollow-sphere volumes of increasing size were generated with $1\,\mathrm{mm}$ isotropic voxel spacing (i) \textbf{small}: $64\times 64\times 64$ voxels; (ii) \textbf{medium}: $96\times 96\times 96$ voxels; (iii) \textbf{large}: $128\times 128\times 128$ voxels. Each contains a hollow sphere with wall attenuation $\approx 1800\,\mathrm{HU}$
and interior $\approx 200\,\mathrm{HU}$. Trilinear interpolation is used during sampling.

\paragraph{Mesh}
One axis-aligned rectangular box implant ($24\times 12\times 50\,\mathrm{mm}$, 12 triangles, scalar value $2200\,\mathrm{HU}$, solid mode) is used in all mesh and hybrid scenes. Both ray-intersection backends are benchmarked separately (see Section~\ref{ssec:backends-comparison}).

\subsubsection{Projection geometry and quality profiles}

Cone-beam geometry: source at $(0,\;0,\;-220)\,\mathrm{mm}$, detector centre at $(0,\;0,\;180)\,\mathrm{mm}$ (source-to-detector distance $400\,\mathrm{mm}$). Physics: Beer--Lambert attenuation integral, $\mu_{\mathrm{water}}=0.02\,\mathrm{mm}^{-1}$.

\begin{table}[!ht]
\centering
\caption{Quality profile parameters used in the benchmark.}
\label{tab:profiles}
\small
\begin{tabular}{lccc}
\toprule
Profile & Detector & Pixel size & Marching step \\
\midrule
\textbf{draft}  & $256\times 256$ & $0.8\,\mathrm{mm}$ & $2.0\,\mathrm{mm}$ \\
\textbf{normal} & $512\times 512$ & $0.4\,\mathrm{mm}$ & $1.0\,\mathrm{mm}$ \\
\textbf{high}   & $512\times 512$ & $0.4\,\mathrm{mm}$ & $0.5\,\mathrm{mm}$ \\
\bottomrule
\end{tabular}
\end{table}

\subsubsection{Performance results}
Table~\ref{tab:main} presents total projection time and throughput per scene for all quality profiles. Table~\ref{tab:phases_master} provides a unified phase breakdown for the normal profile, isolating the primary computational bottlenecks.

\begin{table}[!ht]
\centering
\caption{Total projection time and throughput per scene and quality profile. Time is given in ms (or s if $\geq 1\,\mathrm{s}$). Throughput metrics: \emph{samp/s} denotes attenuation samples (interpolations) per second; \emph{rays/s} denotes fully-traversed rays per second.}
\label{tab:main}
\scriptsize
\begin{tabular}{l rrr rrr rrr}
\toprule
  & \multicolumn{3}{c}{\textbf{draft}}
  & \multicolumn{3}{c}{\textbf{normal}}
  & \multicolumn{3}{c}{\textbf{high}} \\
\cmidrule(lr){2-4}\cmidrule(lr){5-7}\cmidrule(lr){8-10}
Scene \& Backend & time & samp/s & rays/s & time & samp/s & rays/s & time & samp/s & rays/s \\
\midrule
  sampling & & & & & & & & & \\
\midrule
  vol small ($64^3$) & 40\,ms & 12.5M & 509.7k & 335\,ms & 12.1M & 244.3k & 579\,ms & 13.9M & 141.2k \\
  vol medium ($96^3$) & 101\,ms & 15.4M & 460.2k & 1.15\,s & 10.9M & 161.9k & 2.26\,s & 11.1M & 82.7k \\
  vol large ($128^3$) & 241\,ms & 13.6M & 271.7k & 2.50\,s & 10.5M & 104.8k & 4.90\,s & 10.8M & 53.5k \\
\midrule
  vol small + mesh & 304\,ms & 2.6M & 85.7k & 1.51\,s & 4.2M & 69.1k & 1.94\,s & 6.6M & 53.9k \\
  vol medium + mesh & 470\,ms & 4.8M & 126.3k & 2.84\,s & 6.3M & 83.4k & 4.34\,s & 8.2M & 54.7k \\
  vol large + mesh & 636\,ms & 6.5M & 103.0k & 4.39\,s & 7.5M & 59.7k & 7.32\,s & 9.0M & 35.8k \\
\midrule
  Siddon & & & & & & & & & \\
\midrule
  vol small ($64^3$) & 230\,ms & --- & 88.8k & 857\,ms & --- & 95.4k & 898\,ms & --- & 91.1k \\
  vol medium ($96^3$) & 994\,ms & --- & 46.9k & 3.93\,s & --- & 47.5k & 3.94\,s & --- & 47.3k \\
  vol large ($128^3$) & 1.96\,s & --- & 33.4k & 7.85\,s & --- & 33.4k & 7.76\,s & --- & 33.8k \\
\midrule
  vol small + mesh & 529\,ms & 56.6k & 49.3k & 2.14\,s & 56.0k & 48.8k & 2.16\,s & 55.3k & 48.2k \\
  vol medium + mesh & 1.56\,s & 40.5k & 38.0k & 6.08\,s & 41.5k & 39.0k & 6.12\,s & 41.2k & 38.7k \\
  vol large + mesh & 2.31\,s & 30.1k & 28.4k & 9.10\,s & 30.5k & 28.8k & 9.12\,s & 30.4k & 28.7k \\
\midrule
  \texttt{analytic\_bvh} & & & & & & & & & \\
\midrule
  mesh only & 231\,ms & 16.5k & 8.3k & 891\,ms & 17.1k & 8.6k & 877\,ms & 17.4k & 8.8k \\
\bottomrule
\end{tabular}
\end{table}

\begin{table}[!ht]
\centering
\caption{Unified phase breakdown [ms] for the \textbf{normal} profile across all scene configurations. Minor phases (\emph{Depth clipping} and \emph{Physics conversion}) consistently required $<\!1$\,ms and are omitted for space; percentages indicate relative share of total runtime.}
\label{tab:phases_master}
\scriptsize
\begin{tabular}{ll rrrr r}
\toprule
Context & Scene Size & Ray setup & AABB Intersect. & Mesh BVH & Vol. Marching & \textbf{Total} \\
\midrule
\textbf{Volumetric}  & \texttt{small}  & 23 (7\%)  & 20 (6\%)  & $<\!1$      & 292 (87\%)   & \textbf{335} \\
(Sampling)                & \texttt{medium} & 25 (2\%)  & 15 (1\%)  & $<\!1$      & 1111 (96\%)  & \textbf{1153} \\
                          & \texttt{large}  & 26 (1\%)  & 15 (1\%)  & $<\!1$      & 2461 (98\%)  & \textbf{2503} \\
\midrule
\textbf{Volumetric}  & \texttt{small}  & 23 (3\%)  & 16 (2\%)  & 819 (96\%)  & $<\!1$       & \textbf{857} \\
(Siddon)                  & \texttt{medium} & 26 (1\%)  & 16 (0\%)  & 3886 (99\%) & $<\!1$       & \textbf{3929} \\
                          & \texttt{large}  & 22 (0\%)  & 16 (0\%)  & 7807 (100\%)& $<\!1$       & \textbf{7846} \\
\midrule
\textbf{Hybrid}& \texttt{small}  & 23 (1\%)  & 16 (1\%)  & 1044 (69\%) & 428 (28\%)   & \textbf{1511} \\
Vol+Mesh                & \texttt{medium} & 23 (1\%)  & 15 (1\%)  & 1327 (47\%) & 1479 (52\%)  & \textbf{2844} \\
(Sampling)                & \texttt{large}  & 25 (1\%)  & 17 (0\%)  & 1377 (31\%) & 2970 (68\%)  & \textbf{4389} \\
\midrule
\textbf{Hybrid}& \texttt{small}  & 25 (1\%)  & 20 (1\%)  & 2092 (98\%) & $<\!1$       & \textbf{2137} \\
Vol+Mesh                  & \texttt{medium} & 22 (0\%)  & 18 (0\%)  & 6041 (99\%) & $<\!1$       & \textbf{6081} \\
(Siddon)                  & \texttt{large}  & 23 (0\%)  & 16 (0\%)  & 9060 (100\%)& $<\!1$       & \textbf{9100} \\
\midrule
\textbf{Mesh Only}        & \texttt{mesh\_only} & 21 (2\%) & 20 (2\%)  & 850 (95\%)  & $<\!1$       & \textbf{891} \\
\bottomrule
\end{tabular}
\end{table}

\subsubsection{Volume backend comparison: sampling vs.~Siddon}\label{ssec:vol_backends}
Two volumetric integration backends were compared: Joseph-type sampling, based on uniform ray marching with step size \(\Delta s\) and trilinear interpolation, and Siddon voxel traversal, based on analytical voxel-boundary crossings and chord-length accumulation.
Total execution times and scaling metrics are tracked under the volume-only scenarios in Table~\ref{tab:main}.

\begin{figure}[!ht]
\centering
\includegraphics[width=\linewidth]{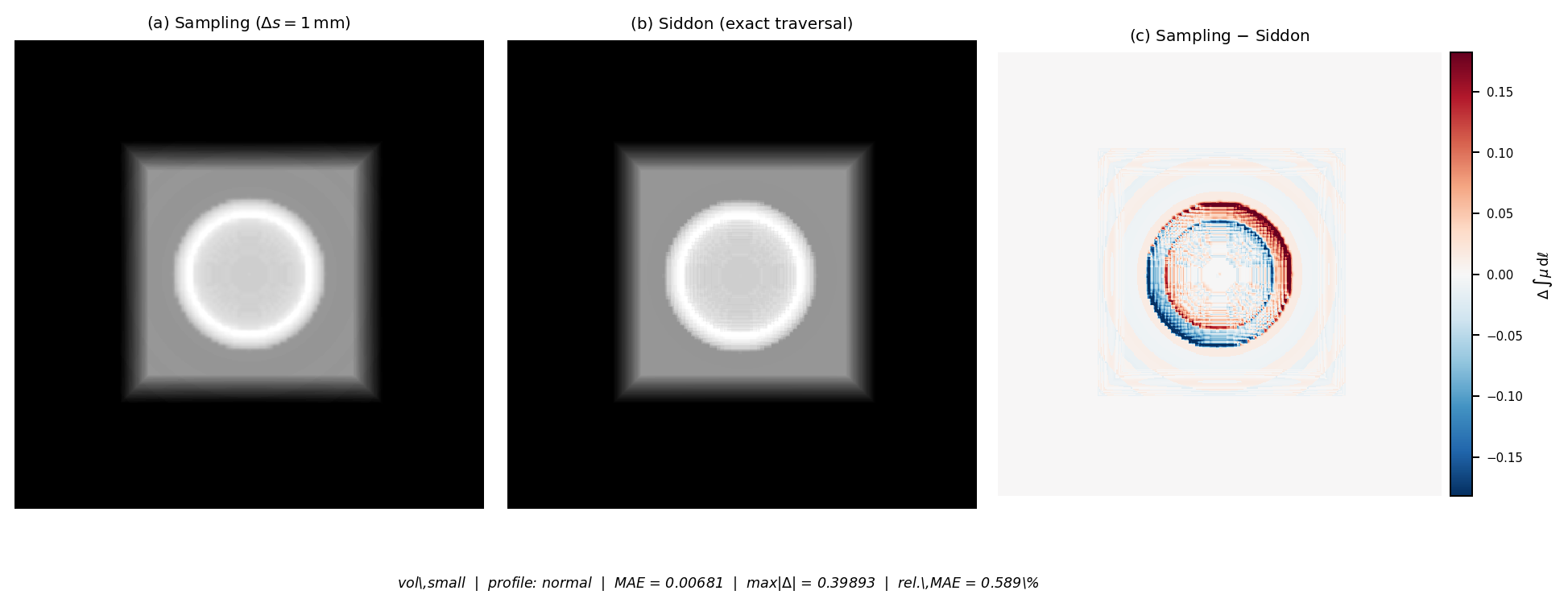}
\caption{Comparison of synthetic cone-beam projections generated using Joseph-type sampling ($\Delta s = 1\,\mathrm{mm}$) and Siddon exact voxel-traversal backends for the \texttt{vol,small} phantom ($64^3$, normal profile, $512\times512$ px). Panel~(a): sampling result. Panel~(b): Siddon result. Panel~(c): signed difference map (sampling -- Siddon), where red/blue denote local deviations.}
\label{fig:sample_siddon}
\end{figure}

Figure~\ref{fig:sample_siddon} illustrates the influence of the volumetric integration backend on the synthetic cone-beam projection under identical geometric and acquisition setups. 

As expected, deviations are concentrated near high-contrast boundaries where Siddon computes exact boundary intersections while the sampling backend approximates the integral via ray-marching. Globally, their discrepancy remains limited with a relative MAE below $1.2\%$ across all configurations (Table~\ref{tab:diff_stats}). Increasing sampling density systematically minimizes this error. Consequently, the sampling backend offers a highly efficient compromise for routine generation. For the tested datasets, relative differences remained below 1.2\%, while execution times were approximately three times shorter than for Siddon traversal. Siddon therefore serves primarily as a step-independent reference formulation.

\begin{table}[!ht]
\centering
\caption{Numerical difference (sampling $-$ Siddon) for volume-only scenes evaluated on raw line-integral images.}
\label{tab:diff_stats}
\small
\begin{tabular}{llrrr}
\toprule
Volume & Profile & MAE & max$|\Delta|$ & Relative MAE (\%) \\
\midrule
  \texttt{vol small}  & draft  & $0.00857$ & $0.38076$ & $0.739$ \\
                      & normal & $0.00681$ & $0.39893$ & $0.589$ \\
                      & high   & $0.00572$ & $0.39890$ & $0.496$ \\
  \midrule
  \texttt{vol medium} & draft  & $0.01734$ & $0.40237$ & $1.093$ \\
                      & normal & $0.01382$ & $0.42005$ & $0.873$ \\
                      & high   & $0.01218$ & $0.40979$ & $0.771$ \\
  \midrule
  \texttt{vol large}  & draft  & $0.02736$ & $0.47403$ & $1.171$ \\
                      & normal & $0.02105$ & $0.45935$ & $0.900$ \\
                      & high   & $0.01843$ & $0.46828$ & $0.788$ \\
\bottomrule
\end{tabular}
\end{table}

\subsubsection{Surface backend comparison}\label{ssec:backends-comparison}
To evaluate computational trade-offs between mesh projection strategies, both the \texttt{analytic\_bvh} loop and the accelerated \texttt{projected\_intersection\_list} backends were benchmarked (Table~\ref{tab:backends}).

\begin{table}[!ht]
\centering
\caption{Mesh-only projection performance for both surface backends (throughput in k\,samp/s).}
\label{tab:backends}
\small
\begin{tabular}{l rr rr rr}
\toprule
  & \multicolumn{2}{c}{\textbf{draft}}
  & \multicolumn{2}{c}{\textbf{normal}}
  & \multicolumn{2}{c}{\textbf{high}} \\
\cmidrule(lr){2-3}\cmidrule(lr){4-5}\cmidrule(lr){6-7}
Backend & time & k\,s/s & time & k\,s/s & time & k\,s/s \\
\midrule
  \texttt{analytic\_bvh}                & 231\,ms & 16.5 & 891\,ms & 17.1 & 877\,ms & 17.4 \\
  \texttt{projected\_intersection\_list}& 83\,ms  & 45.6 & 255\,ms & 59.3 & 276\,ms & 54.8 \\
\bottomrule
\end{tabular}
\end{table}

\subsubsection{Projection images}

\begin{figure}[!ht]
\centering
\includegraphics[width=.8\linewidth]{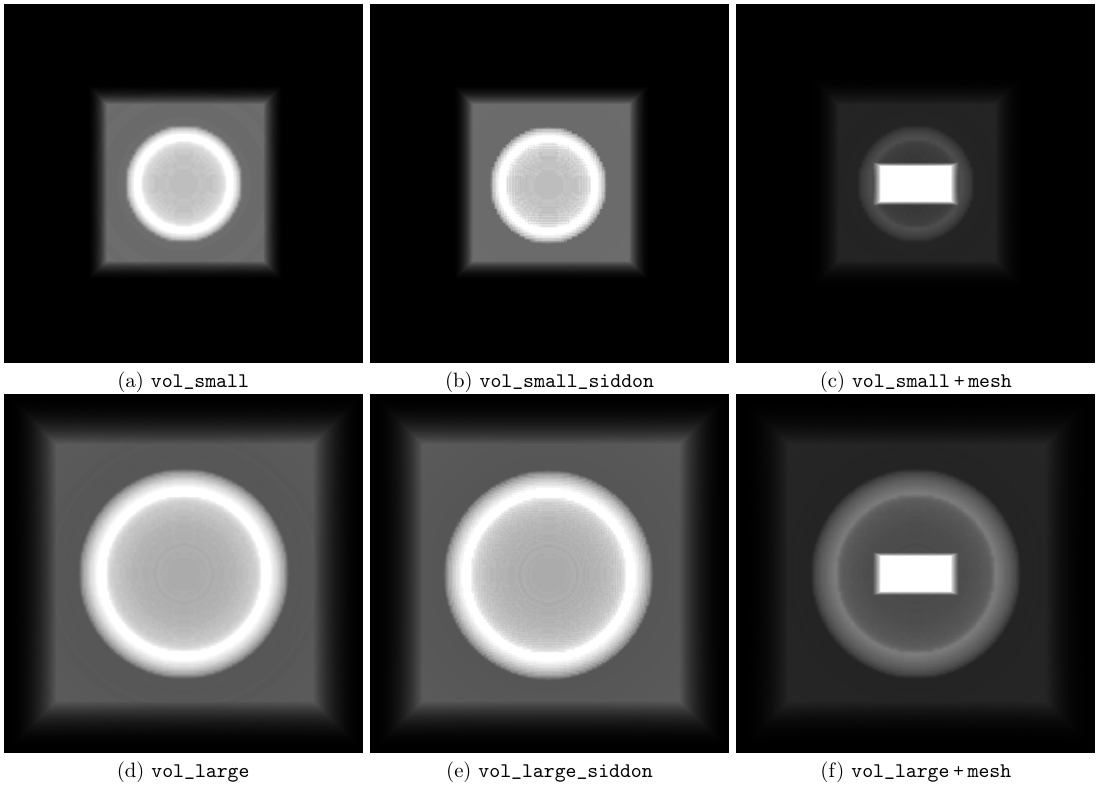}
\caption{Synthetic cone-beam projections, normal profile
  ($512\times512$\,px, step\,$1.0\,\mathrm{mm}$).
  Columns: sampling / Siddon / sampling\,+\,mesh.
  Rows: small ($64^3$)
  / large ($128^3$) volume.}
\label{fig:proj_normal}
\end{figure}

Figure~\ref{fig:proj_normal} shows synthetic cone-beam projections for 
selected
scene variants
at the \textbf{normal} quality profile ($512\times512$\,px, step\,$1.0\,\mathrm{mm}$),
rendered with a digital radiography presentation model
(standard convention: dense~=~white, $\gamma=0.6$, contrast\,=\,1.1).
Images at the draft and high profiles are visually indistinguishable
for the synthetic objects used here.

\FloatBarrier

\subsection{Multimodal anatomical scene projection and transformation-driven scenarios}

The examples presented in this section are intended to illustrate the
capabilities of the proposed transformation-driven framework when applied to
heterogeneous anatomical data. The objective is not quantitative assessment of
radiographic fidelity or diagnostic equivalence with clinical radiographs, but
demonstration of multimodal scene integration, transformation-aware projection
generation, anatomical annotation propagation, and motion-related projection
scenarios.

Real-data experiments were therefore designed to demonstrate flexibility of
scene representation rather than quantitative radiographic fidelity.

A series of illustrative experiments was performed using real CBCT datasets
together with surface models obtained from optical scanners and from surface
reconstructions derived from volumetric data.

\subsubsection{Projection of multiple anatomical models}

\begin{figure}[hb!]
    \centering\scriptsize
    \includegraphics[width=\linewidth]{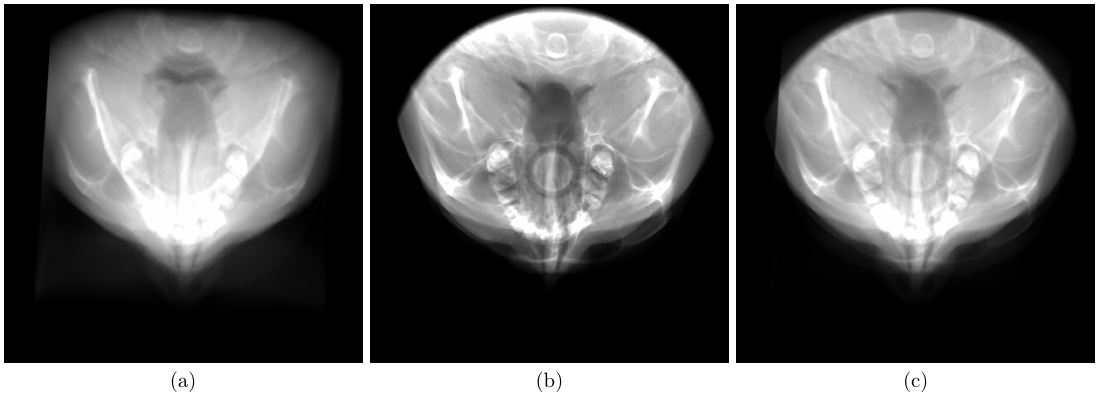}
    \caption{
Synthetic projections generated from two CBCT datasets in the superior--inferior
viewing direction. Panels (a) and (b) show projections of the individual
volumes, whereas panel (c) shows the projection of the registered combined
scene.
}
    \label{fig:2cbct}
\end{figure}

Figure~\ref{fig:2cbct} presents synthetic projections generated from two CBCT examinations. The data consisted of anonymised retrospective datasets used exclusively for methodological demonstration, and no diagnostic conclusions were derived from them. The datasets differed in spatial resolution (voxel size: 0.288\,mm and 0.200\,mm, respectively) and partially covered different anatomical regions. After registration, both volumes were projected individually and as a combined multimodal scene.

The example demonstrates that independently acquired volumetric datasets may be integrated within a common spatial reference frame and subsequently visualised through a single projection process. Such functionality may be useful in situations where several incomplete datasets provide complementary anatomical information. Because projection generation operates directly on the scene representation, the individual datasets do not need to be merged into a common resampled volume prior to visualisation.

The projections were generated along a superior--inferior viewing direction, with the source positioned below the head and the detector above it. Such a configuration is generally unavailable in conventional radiographic acquisition and illustrates the flexibility of the virtual projection environment.

\begin{figure}[h!]
    \centering\scriptsize
    \includegraphics[width=\linewidth]{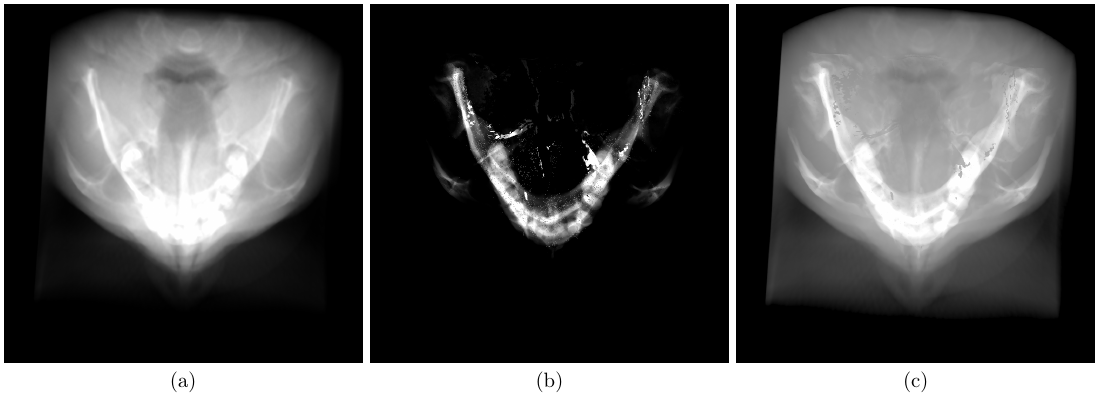}
    \caption{
Synthetic projections generated from a CBCT dataset and a surface model in the
superior--inferior viewing direction. Panel (a) shows the projection of the
CBCT volume, panel (b) shows the projection of the surface model, and panel
(c) shows the projection of the combined volumetric--surface scene.
}
    \label{fig:cbctrekonstr}
\end{figure}

Figure~\ref{fig:cbctrekonstr} presents a related experiment in which a CBCT volume is combined with a surface reconstruction of the craniofacial skeleton. This example illustrates the ability of the framework to jointly process volumetric and surface-based anatomical representations within a common projection pipeline. At the same time, it highlights limitations associated with mesh-based projection, where inaccuracies of the triangular representation or topological imperfections may introduce visible projection artefacts.

The computational cost of such multimodal projections depends on the selected projection backend, acquisition parameters, and scene complexity. For surface-based models, execution time additionally depends on mesh complexity, particularly on the number of triangles participating in the projection process. When multiple objects are projected simultaneously, individual computational costs accumulate, potentially increasing execution times from seconds to minutes on less powerful hardware.

\begin{figure}[h!]
    \centering
    \includegraphics[width=\linewidth]{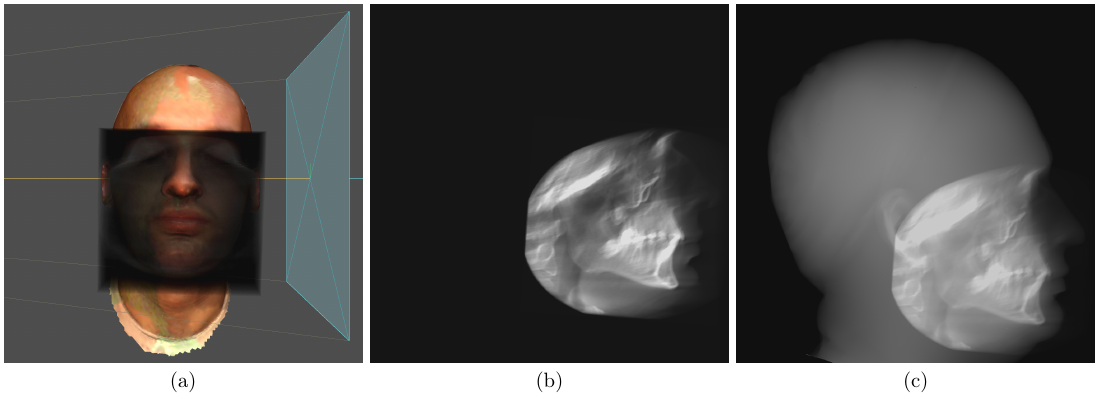}
    \caption{
Multimodal projection example illustrating the integration of heterogeneous
anatomical data within a common VirtualRTG scene. Panel (a) shows several
registered volumetric and surface-based objects positioned within the
projection geometry, panel (b) shows the projection generated from the CBCT
volume alone, and panel (c) shows the projection after adding surface-based
multimodal data, illustrating how complementary models may enrich the
projection-space representation.
}
    \label{fig:multimodalRTG}
\end{figure}

The ability to combine heterogeneous anatomical representations is further illustrated in Figure~\ref{fig:multimodalRTG}, where a CBCT dataset is projected together with additional multimodal information. In such scenarios, balancing the visual contribution of individual objects may become challenging, since highly attenuating structures can dominate the resulting image. To address this issue, the framework allows independent assignment of attenuation properties to individual surface models.

From a methodological perspective, these examples demonstrate that projection imaging may be applied not only to a single volumetric acquisition but also to complex multimodal anatomical scenes containing data originating from different imaging modalities. Such integration may potentially reduce the extent of volumetric acquisition required for a given task by complementing selected anatomical structures with information obtained from surface-based imaging techniques.

\subsubsection{Region-of-interest selection (slab) and temporomandibular-joint visualisation}

The slab mechanism enables attenuation accumulation to be restricted to a selected spatial interval without modifying the underlying anatomical model. As a result, projection formation may be focused on specific anatomical regions while suppressing structures located outside the selected depth range.

\begin{figure}[h!]
    \centering
    \includegraphics[width=\linewidth]{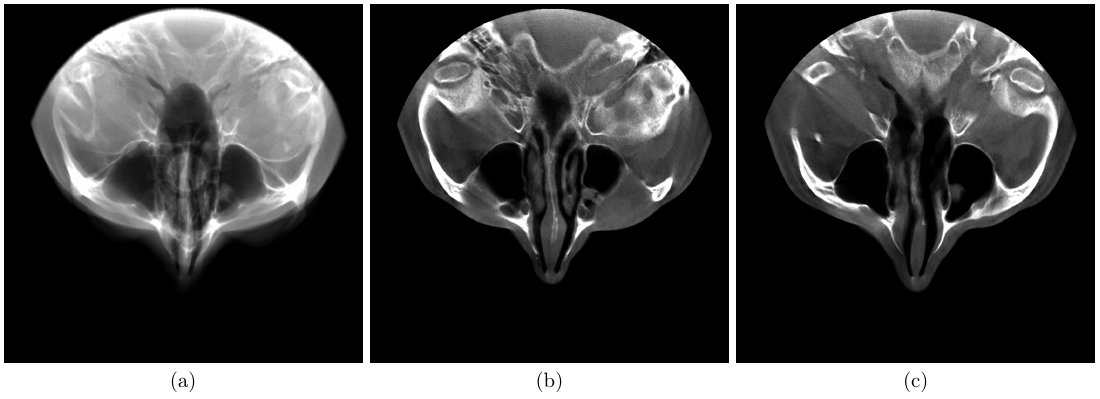}
    \caption{
Depth-limited synthetic projections used for temporomandibular-joint
visualisation. Panel (a) shows an axial projection generated from a broad
\(35\,\mathrm{mm}\) slab, while panels (b) and (c) show projections generated
from narrower \(5\,\mathrm{mm}\) slabs positioned to visualise the right and
left temporomandibular joints, respectively.
}
    \label{fig:stawosiowo}
\end{figure}

Figure~\ref{fig:stawosiowo} demonstrates this concept for visualisation of the temporomandibular joints. By reducing the active projection thickness from a broad anatomical region to a narrow slab, it becomes possible to isolate selected joint structures and improve the visibility of the articular space. Depending on anatomical asymmetry and object positioning, several slab positions may be required to obtain optimal visualisation of the left and right joints.

More generally, the slab mechanism illustrates how prior knowledge of three-dimensional anatomy may be used to generate selective projection images that are not directly achievable in conventional radiography. For example, projection formation may be restricted to selected anatomical regions, including only one side of the craniofacial complex. Because conventional radiographs represent a superposition of structures originating from different depths, selective projection may reduce ambiguities associated with overlapping anatomy and facilitate interpretation of specific regions of interest.

Such projections should not be interpreted as replacements for standard radiographic examinations. Rather, they constitute an exploratory visualisation tool that enables analysis of relationships between three-dimensional anatomy and its projection-space representation.

\subsubsection{Projection of anatomical landmarks}

\begin{figure}[b!]
    \centering
    \includegraphics[width=\linewidth]{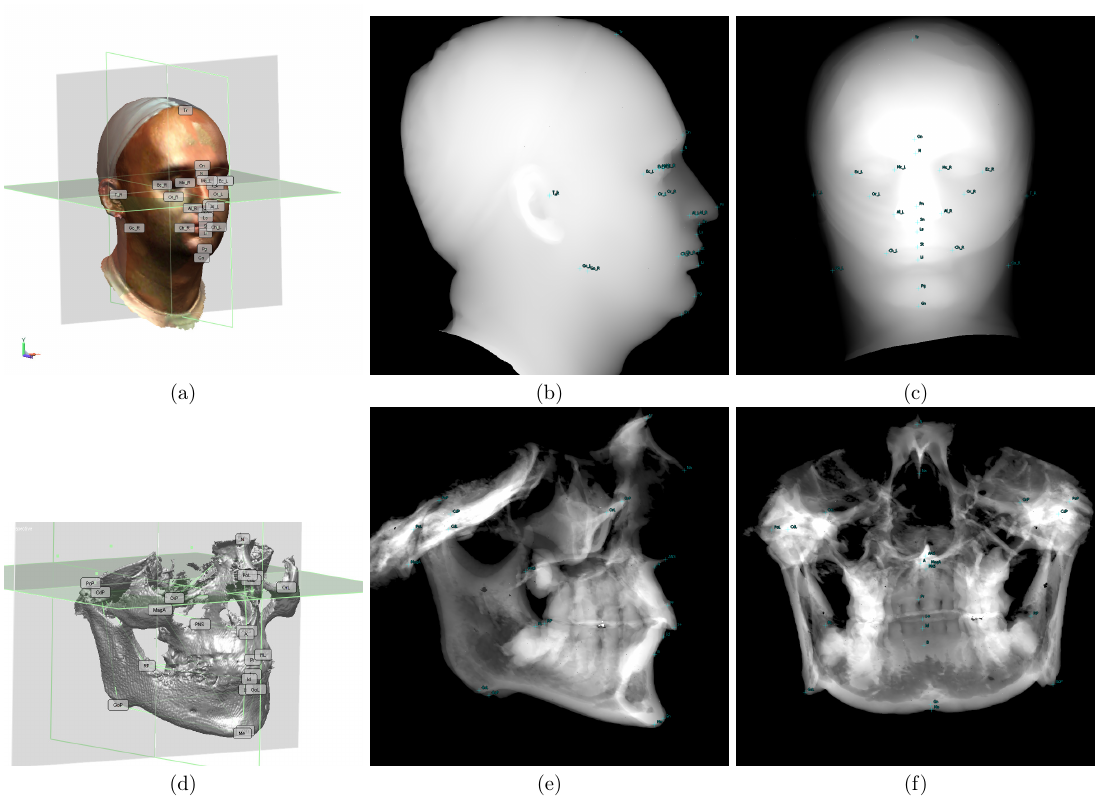}
    \caption{
Projection of anatomical landmarks associated with different anatomical
representations. Panel (a) shows a three-dimensional facial surface model
with visible landmarks, whereas panels (b) and (c) show the corresponding
lateral cephalometric and posteroanterior projections. Panel (d) presents
a skeletal surface reconstruction with anatomical landmarks, while panels
(e) and (f) show the corresponding lateral cephalometric and
posteroanterior projections. The example illustrates that annotations
associated with different anatomical representations participate in the
same transformation-aware scene and can be propagated consistently into
projection space.
}
    \label{fig:punkty}
\end{figure}

Anatomical landmarks constitute a fundamental component of cephalometric
analysis and have been extensively used in orthodontic diagnostics for
several decades. Figure~\ref{fig:punkty} presents examples of landmark
projection for different anatomical representations.
Landmarks defined on
a three-dimensional facial surface model as well as landmarks associated
with a skeletal surface reconstruction are propagated through the same
transformation-aware scene and projected into image space together with
their corresponding anatomical structures. The resulting VirtualRTG projections preserve the spatial relationships
defined within the multimodal scene while enabling direct visualisation of
anatomical annotations in projection space.

Because landmarks participate in the same transformation hierarchy as anatomical objects, their projected positions remain consistent under changes of projection geometry and anatomical configuration. This enables direct correspondence between anatomical annotations and
measurements defined in three-dimensional space and their projection-space representations.

The presented example demonstrates that projection-space annotations may serve as a bridge between traditional two-dimensional cephalometric workflows and modern three-dimensional anatomical analysis. Potential applications include validation of cephalometric measurements, comparison between 2D and 3D approaches, and generation of annotated projection datasets for methodological studies.

\FloatBarrier
\subsubsection{Dynamic anatomical models and motion analysis}

\begin{figure}[ht!]
    \centering
    \includegraphics[width=\linewidth]{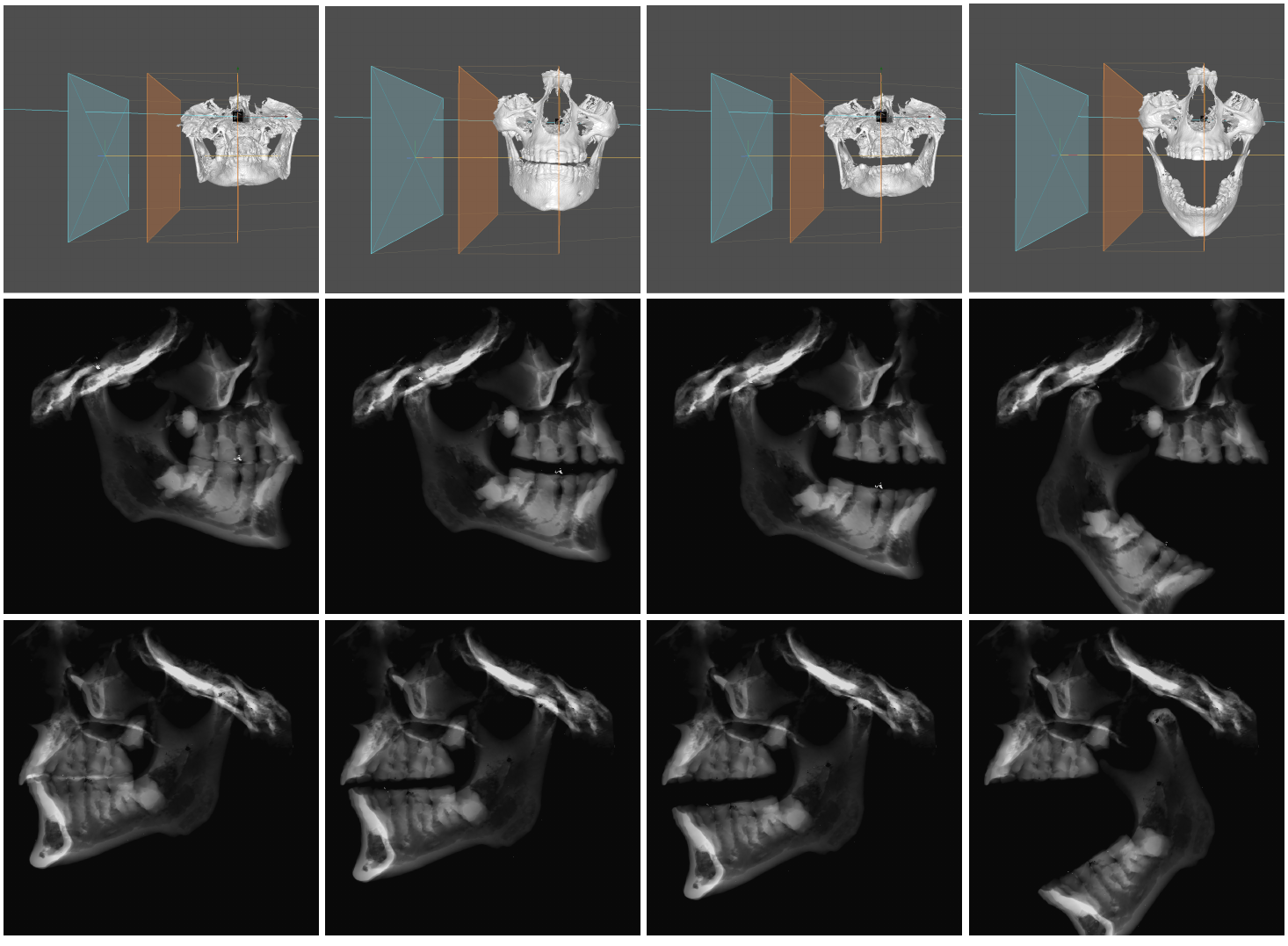}
    \caption{
Transformation-driven projection imaging of mandibular opening using
depth-limited half-side projections. Columns show consecutive configurations
selected from the recorded opening motion. The top row presents schematic
three-dimensional views of the model and projection geometry, while the middle
and bottom rows show slab-based synthetic projections generated separately for
the left and right sides. Separate depth-limited projections reduce overlap of
contralateral anatomical structures and improve visualisation of the
temporomandibular-joint region during motion.
}
    \label{fig:ruch}
\end{figure}

Figure~\ref{fig:ruch} presents an example in which anatomical motion is propagated directly into projection space through scene transformations. The example is based on multimodal anatomical data previously used in motion-analysis studies and illustrates four selected phases of mouth
opening.
In conventional radiography, observation of such motion would require repeated
image acquisition at multiple jaw positions. In contrast, the proposed
framework enables generation of comparable projection images from a single
registered anatomical scene, allowing projection-space effects of mandibular
motion to be explored under identical acquisition conditions.

For each configuration, synthetic projections of the temporomandibular joints
were generated under controlled geometric conditions. The projections were
computed as depth-limited half-side views, with separate slab settings used for
the left and right sides in order to reduce superposition of contralateral
anatomical structures. To preserve comparable magnification and viewing
geometry, the anatomical model was rotated between left- and right-side
projections, allowing both joints to be analysed under equivalent projection
conditions.

In the current implementation, projections are generated from surface reconstructions assigned a uniform attenuation coefficient. Consequently, image quality is influenced by the accuracy of the underlying triangular meshes. Geometric artefacts may arise from limited surface approximation accuracy as well as from mesh-topology defects. Despite these limitations, the resulting projections preserve sufficient information to visualise condylar displacement and changes in joint relationships during mandibular motion.

From a methodological perspective, this example illustrates how
anatomical motion can be propagated directly into projection space while
preserving identical acquisition geometry and imaging assumptions.
Consequently, multiple directly comparable projection images may be
generated from a shared anatomical reference scene.
Future developments may further improve projection quality by separating
movable and static anatomical components directly within volumetric data
and performing motion simulation without intermediate surface
reconstruction.

\subsection{Key methodological observations}

The conducted experiments support the four methodological assumptions underlying the proposed framework. First, the real-data scenarios demonstrate that heterogeneous anatomical objects can be projected within a common transformation-aware scene without mandatory resampling into a single volume. Registered CBCT datasets, reconstructed surfaces, optical scans, landmarks, and auxiliary geometric objects may therefore remain represented as independent scene entities while contributing jointly to the resulting projection-space representation.

Second, the synthetic benchmarks identify the dominant computational components of the current implementation. In volume-only scenarios, projection time is governed primarily by volumetric ray marching or voxel traversal. In mesh-based and hybrid scenarios, ray--surface intersection becomes a major bottleneck, and the total cost increases with the number and complexity of objects participating in the projection process.

Third, the comparison between Joseph-type sampling and Siddon traversal shows the expected trade-off between computational efficiency and step-size-independent voxel traversal. For the tested synthetic scenes, numerical differences between the two approaches remained limited and were concentrated mainly near high-contrast boundaries. This indicates that sampling provides a practical backend for routine projection generation, whereas Siddon may serve as a reference formulation when exact voxel traversal is required.
The phase breakdown further indicates that scene-management overhead remains small. Ray initialisation, coordinate transformations, and bounding-volume preparation account for only a minor fraction of the total runtime, whereas more than 90\% of the execution time is consistently spent in projection-specific computations, namely volumetric integration or ray--surface intersection processing.

Fourth, the depth-limited and motion-related examples demonstrate that transformation-aware projection generation can be used to analyse selected anatomical relationships under controlled conditions. Slab-based projections reduce structural superposition in temporomandibular-joint visualisation, while mandibular-motion examples show that multiple directly comparable VirtualRTG projections can be generated from a shared anatomical reference scene while preserving identical acquisition and presentation assumptions.

Overall, the presented experiments indicate that the main value of the framework lies not in reproducing clinical radiographs with maximum physical realism, but in providing a controllable computational environment for studying relationships between anatomy, motion, multimodal data, and their projection-space representations.

\FloatBarrier
\section{Discussion and framework limitations}

The proposed framework should be interpreted primarily as a methodological and computational environment for controlled experimentation rather than a fully physically faithful simulator of clinical radiographic acquisition. Its design prioritises explicit representation of anatomical transformations, controlled projection geometry, reproducibility of experimental conditions, and generation of multiple comparable synthetic projections from a shared anatomical scene.

From this perspective, several modelling decisions adopted in the
present work should be interpreted as deliberate modelling
compromises rather than missing components. The framework prioritises controllability, interpretability, architectural flexibility, and compatibility with transformation-driven anatomical modelling over maximal physical realism.

The framework assumes that all anatomical objects participating in projection
formation are already represented within a common spatial reference frame.
Registration, segmentation, and surface reconstruction are therefore treated as
external preprocessing steps and are not the focus of the present work. Errors
introduced during these steps, including residual registration inaccuracies,
segmentation errors, or mesh-reconstruction artefacts, propagate directly to the
generated projections and may affect their anatomical interpretation.

The reported performance figures correspond to a CPU-based Python implementation and should therefore be interpreted as representative of the proposed framework rather than of the underlying projection algorithms themselves. The present work focuses on scene representation, transformation management, and projection reproducibility. Hardware-specific acceleration, including GPU-based implementations, remains outside the current scope and constitutes a straightforward direction for future development.

An important advantage of the proposed formulation is that it enables
investigation of hypothetical or therapy-related anatomical configurations that
may not be directly available through clinical imaging. Examples include
mandibular repositioning, splint-assisted treatment scenarios, or controlled
kinematic studies of TMJ-related motion.

In such scenarios, repeated acquisition of radiographic images at multiple
anatomical configurations would be impractical, difficult to standardise, or
associated with additional patient radiation exposure. By deriving multiple
projection images from a common anatomical reference scene, the proposed
framework enables systematic exploration of projection-space consequences of
anatomical transformations while preserving identical acquisition geometry and
imaging assumptions.

\subsection{Scope of radiographic interpretation}
The generated projections should be interpreted as controlled synthetic observations of explicitly represented anatomical scenes rather than as diagnostically equivalent substitutes for clinical radiographs. The present implementation intentionally uses simplified attenuation and detector models and does not account for polychromatic spectra, photon scattering, focal-spot effects, detector-specific response, or quantum noise. Moreover, CBCT voxel values are not universally calibrated attenuation coefficients and may depend on scanner type, acquisition protocol, reconstruction settings, and artefacts. Consequently, the proposed framework is intended for methodological experimentation, analysis of anatomy--projection relationships, and generation of comparable transformation-dependent scenarios, rather than for direct clinical diagnosis or quantitative radiographic dose/contrast prediction.

The following subsections discuss these limitations in more detail, focusing on physical realism, input-data assumptions, computational performance, and future extensions of the framework.

\subsection{Physical simplifications and input-data limitations}

The physical simplifications outlined above have direct consequences for the interpretation of the generated projections. The resulting images preserve selected geometric and attenuation-related properties under controlled assumptions, but should not be interpreted as physically faithful replicas of clinical radiographs.

An important limitation arises from the underlying CT/CBCT data themselves. Although CT intensities may approximately relate to attenuation properties through Hounsfield units (HU), scanner-specific reconstruction procedures, acquisition settings, and preprocessing limit direct comparability across datasets. The problem is more pronounced in CBCT, where scalar values are not universally calibrated and may vary substantially between devices, acquisition protocols, and reconstruction pipelines. Consequently, the framework should not be interpreted as a tool for quantitative estimation of physical attenuation from CBCT values. Instead, it provides a configurable environment for investigating how alternative material-interpretation assumptions influence projection-space representations.

As a consequence, uncertainty present in volumetric input data propagates to material-response modelling and final image formation. For this reason, the framework intentionally avoids assuming a single fixed interpretation of voxel intensities. Instead, preprocessing, material interpretation, attenuation weighting, and presentation remain explicitly separated and configurable, allowing alternative modelling assumptions to be explored without modifying the remainder of the acquisition pipeline.

\subsection{Geometric and computational trade-offs}

The projection engine primarily relies on ray marching with interpolated sampling rather than exact voxel-intersection schemes. This choice reflects a trade-off between geometric exactness, computational cost, and compatibility with independently transformable anatomical objects embedded within a hierarchical scene graph.

Unlike reconstruction-oriented voxel traversal methods, the adopted approach naturally supports dynamically transformed objects, heterogeneous anatomical representations, and projection of multiple independently positioned volumes without prior resampling into a common voxel grid. At the same time, numerical accuracy depends on ray-sampling density and interpolation strategy. Larger sampling steps reduce computational cost but may reduce sensitivity to fine anatomical detail.

The approximate computational complexity may be expressed as:
\[ T \sim H \cdot W \cdot \frac{L}{\Delta s} \cdot N_v, \]
where \(H \cdot W\) denotes detector resolution, \(L\) the average ray length, and \(N_v\) the number of active volumetric sources.

The framework therefore exposes configurable quality profiles controlling sampling density and detector resolution, enabling practical compromises between execution time and image quality during exploratory and comparative experiments.

An additional limitation concerns modalities involving dynamically changing acquisition geometry. While cephalometric and cone-beam-inspired projections fit naturally within a fixed scene geometry, modalities such as panoramic imaging require moving acquisition trajectories and image-composition logic extending beyond a single static projection model.

\subsection{Methodological implications and future directions}

The methodological value of the framework stems primarily from its
transformation-driven formulation and controlled observability rather than
maximal physical realism. In the proposed approach, anatomical transformations
constitute explicit entities within the scene representation and are propagated
directly into projection space. Consequently, the framework enables systematic
investigation of how anatomically defined transformations manifest in projection
images without requiring repeated image acquisition for every analysed
configuration.

Because anatomical transformations, projection geometry, material
interpretation, and presentation remain explicitly parameterised, projection
changes may be analysed under fully reproducible conditions. This creates a
controlled experimental environment for studying projection-space consequences
of anatomical motion, therapeutic repositioning, and other transformation-driven
scenarios.

A particularly important application concerns anatomical structures whose
behaviour changes over time, such as the temporomandibular joint (TMJ) during
mandibular motion. A shared anatomical reference scene may serve as a basis for
multiple alternative configurations, enabling directly comparable synthetic
projections corresponding to different motion phases or treatment-related
positions without repeated radiographic exposure. Such a scene may be derived
from a single CT/CBCT acquisition, from segmented volumetric components, or
from multiple registered volumetric and surface-based datasets.

Because anatomical transformations remain explicitly known, the framework may
support generation of annotated projection datasets with reference motion
information, facilitating validation of 2D--3D registration, motion tracking,
geometry-sensitive image analysis, and machine-learning methods based on
projection data.

Beyond motion-related applications, the framework further supports
heterogeneous multimodal scenes in which volumetric and surface-based
anatomical representations coexist within a common transformation hierarchy.
Consequently, CT/CBCT volumes, segmented structures, dental scans, splints,
and auxiliary geometries may contribute jointly to image formation while
preserving explicit spatial relationships.

The framework additionally supports propagation of anatomical annotations from
the three-dimensional scene into projection space. Since landmarks and paths
remain embedded within the same transformation hierarchy as anatomical
structures, their projected positions remain geometrically consistent under
anatomical motion and changing acquisition geometry. This may support
cephalometric analysis, therapeutic planning, motion interpretation, and
generation of annotated synthetic projection datasets.

Future work may focus on extending physical realism, supporting more complex
projection geometries, and incorporating richer anatomical motion models,
including non-rigid deformation and biomechanical constraints. Nevertheless,
the central modelling idea remains unchanged: explicit and reproducible
propagation of anatomical transformations into projection-space
representations.

\section{Conclusions}

This work introduced a transformation-driven framework for synthetic projection imaging in which independently transformable anatomical structures embedded in a scene graph give rise to comparable projection-space representations. Rather than treating projection generation as an isolated rendering task, the proposed approach formulates synthetic radiographic imaging as a downstream consequence of explicitly represented anatomy, spatial transformations, projection geometry, material-response modelling, and presentation.

A central contribution of the framework lies in linking multimodal anatomical representations with a projection-space observation layer. Anatomical configurations originating from registration, treatment planning, therapeutic positioning, or motion modelling may be propagated directly to synthetic projection images while preserving explicit control over acquisition assumptions and scene geometry.

The proposed environment naturally supports transformation-driven anatomical scenarios, particularly those involving mandibular motion and therapeutic configurations. A shared anatomical reference scene may serve as a basis for multiple
alternative spatial configurations, enabling generation of directly comparable
VirtualRTG projections without repeated radiographic acquisition. Depending on
the application, this scene may originate from a single CT/CBCT acquisition,
segmented volumetric components, or multiple registered volumetric and
surface-based datasets.

Beyond conventional volume-only simulation, the framework supports multimodal
projection scenarios in which volumetric and surface-based anatomical objects
coexist within a common transformation hierarchy. Consequently, CT/CBCT data,
segmented structures, dental scans, splints, and auxiliary geometries may
jointly contribute to synthetic image formation while preserving explicit
spatial relationships.

The computational evaluation further demonstrated that the proposed
scene-oriented formulation introduces only a minor management overhead.
Across the analysed benchmark scenarios, execution time was dominated by
projection-specific operations, namely volumetric integration and ray--surface
intersection processing, indicating that hierarchical scene management remains
compatible with practical projection generation.

Rather than aiming at maximal physical realism, the proposed framework
prioritises controlled and reproducible exploration of anatomy--projection
relationships under explicit modelling assumptions.

Future work may focus on improving physical realism, extending projection geometries, incorporating richer anatomical motion models, and developing validation workflows for motion analysis and synthetic projection datasets.


\section*{CRediT authorship contribution statement}
\begin{itemize}
  \item \textbf{Dariusz Pojda}: Conceptualization, Software, Investigation, Project administration, Validation, Visualization, Writing --- original draft, Writing --- review \& editing.
  \item \textbf{Krzysztof Domino}: Supervision, Formal analysis, Validation, Writing --- review \& editing.
  \item \textbf{Michał Tarnawski}: Data curation, Resources, Validation, Writing --- review \& editing.
  \item \textbf{Agnieszka Anna Tomaka}: Conceptualization, Data curation, Formal analysis, Investigation, Methodology, Resources, Validation, Writing --- original draft, Writing --- review \& editing.

\end{itemize}

\section*{Code and data availability}
The source code and scripts required to reproduce the synthetic benchmark experiments are available at \url{https://github.com/iitis/virtRTG.git}. The projection functionality described in this work is implemented as the \texttt{virtRTG} research module for the \texttt{pyDpVision} environment. \texttt{pyDpVision} (\url{https://github.com/pojdulos/pyDpVision.git}) is a Python-based experimental environment derived from the previously published \texttt{dpVision} platform \cite{pojda2025dpvision} and is used here primarily to provide scene hierarchy, object transformation, and projection-workflow management. The released code is intended to support reproducibility of the computational experiments reported in this manuscript, rather than to provide a complete end-user application or a full replacement for the original C++ \texttt{dpVision} software. The real-data demonstration workflow depends on anonymised medical datasets and local preprocessing steps. These datasets cannot be publicly redistributed due to privacy and ethical restrictions.

\section*{Declaration of generative AI use}

During the preparation of this manuscript, the authors used generative
artificial intelligence tools to assist with language editing, text
refinement, and improvement of readability. All scientific content,
methodology, software implementation, experimental design, analysis,
interpretation of results, and final editorial decisions were developed
and verified by the authors. The authors take full responsibility for the
content of this publication.

\bibliographystyle{elsarticle-num}
\bibliography{refs}

\end{document}